\documentclass[runningheads]{llncs}

\usepackage{eccv}

\usepackage{eccvabbrv}

\usepackage{tikz}

\usepackage[utf8]{inputenc} 
\usepackage[T1]{fontenc}    
\usepackage{booktabs}       
\usepackage{amsfonts}       
\usepackage{nicefrac}       
\usepackage{microtype} 
\usepackage{caption}
\usepackage{makecell}
\usepackage{tikz,pgfplots}
\usetikzlibrary{calc}
\usepackage{graphicx, wrapfig}
\usepackage{mathtools}
\usepackage{bm}
\usepackage{array}
\usepackage{multirow}
\usepackage{pifont}
\usepackage{tabularx}
\usetikzlibrary{spy}
\usepackage{longtable}

\newcommand{\vt}[1]{\mathbf{#1}}
\newcommand{\mat}[1]{\bm{#1}}

\newcommand{\surf}[1]{\mathcal{#1}} 
\newcommand{\norm}[1]{\left\| #1 \right\|}
\newcommand{\dotp}[1]{\langle #1 \rangle}

\DeclareMathOperator*{\argmin}{arg\,min}

\newcommand\round[1]{\left[#1\right]}

\newcommand*{\inparagraph}[1]{\noindent\textbf{#1}\hspace{0.5em}}

\definecolor{drawred}{rgb}{0.72, 0,0}
\definecolor{drawblue}{rgb}{0,0.4,0.8}
\usepackage[accsupp]{axessibility}  

\usepackage[pagebackref,breaklinks,colorlinks,citecolor=eccvblue]{hyperref}

\begin{document}

\title{Enhancing Surface Neural Implicits with Curvature-Guided Sampling and Uncertainty-Augmented Representations} 

\titlerunning{Curvature-Guided Uncertainty Surface }

\author{Lu Sang  \and
Abhishek Saroha \and
Maolin Gao
Daniel Cremers}

\authorrunning{S.Lu et al.}

\institute{
Technical University of Munich\and 
Munich Center for Machine Learning \\
\email{lu.sang, abhishek.saroha, maolin.gao, cremers@tum.de}\\
}

\maketitle

\input{figures_tex/teaser.tex}

\begin{abstract}

Neural implicit representations have become a popular choice for modeling surfaces due to their adaptability in resolution and support for complex topology.
While previous works have achieved impressive reconstruction quality by training on ground truth point clouds or meshes, they often do not discuss the data acquisition and ignore the effect of input quality and sampling methods during reconstruction. 
In this paper, we introduce a method that directly digests depth images for the task of high-fidelity 3D reconstruction. 
To this end, a simple sampling strategy is proposed to generate highly effective training data, by incorporating differentiable geometric features computed directly based on the input depth images with only marginal computational cost. 
Due to its simplicity, our sampling strategy can be easily incorporated into diverse popular methods, allowing their training process to be more stable and efficient. 
Despite its simplicity, our method outperforms a range of both classical and learning-based baselines and demonstrates state-of-the-art results in both synthetic and real-world datasets.

\end{abstract}

\section{Introduction}\label{sec:intro}

Reconstructing neural implicit representations of surfaces is crucial for a range of downstream applications~\cite{atzmon2021augmenting, chen2023local}. While numerous approaches utilize point clouds or meshes to model surfaces with Multi-Layer Perceptrons (MLPs), they often assume already available high-quality, dense input. How to acquire these inputs and how to deal with sparse and noisy training data has been often ignored. For example, some methods require (oriented) surface normals to produce satisfactory results~\cite{gropp2020implicit}.
In real-world scenarios, the requirement on high-quality point cloud and its normals is often hard to fulfill, which leads to failure cases of recent reconstruction methods~\cite{gropp2020implicit, chibane2020ndf}. 
Furthermore, the uneven distribution of training data and the influence of sampling strategies on training efficiency and output quality prompt a significant question: How can we tailor the sampling strategy to the characteristics of available input data to improve training? 

In this paper, we introduce an innovative approach that directly leverages raw depth images from sensors to train uncertainty-aware neural implicit functions for high-fidelity 3D reconstruction tasks, ranging from single objects to large scenes. Our uncertainty-aware neural implicits can represent open surfaces, hence extending the representation capacity of SDF, which traditionally requires a clear notion of surface inside and outside.
Our method builds an intermediate representation of a coarse voxel grid with various geometric properties, such as gradients and curvatures, based on input depth images, which remarkably enables continuous sampling inside the (discrete) grid. 
Consequently, our proposed method can handle sparse, low-quality input and reconstruct large open surfaces with high fidelity as shown in Fig.~\ref{fg:teaser}. In summary, our contributions are:

\begin{itemize}
    \item We introduce a novel method, which can deal with raw input depth images to reconstruct surfaces ranging from single objects to large scenes under the same framework. 
    \item We propose a method that computes mean curvature directly from input depth images and devises a simple yet effective sampling strategy to generate training data for faster and more accurate neural implicit fitting.
    \item Our sampling strategy is lightweight in terms of computational time and can be easily integrated into existing neural implicit reconstruction methods.
    \item Our uncertainty-aware implicit neural representation enables, for the first time, SDF-based open surface reconstruction.
\end{itemize}
\section{Related Work}\label{sec:related}
\subsection{Surface Representation}
Surface representation can be classified into two categories based on the stored surface properties: explicit surface representation (\eg, polygon meshes or point clouds) and implicit surface representation (\eg, signed distance fields). Explicit methods struggle with complex topologies, resolution adjustments, local modifications, and potential high memory consumption when storing high-resolution surfaces. In contrast, implicit representations represent the surfaces by storing indirect information about the surface, which overcomes the shortages of explicit methods. However, \textbf{classical} implicit methods still suffer from fixed resolution and high memory consumption issues. For example, the memory consumption is $\mathcal{O}(n^3)$ for a voxel grid of size $n^3$. Luckily, \textbf{neural} implicit representations can encode the surface implicit information such as occupancies~\cite{lars2018occupancy, genova2019Deep,chibane2020implicit}, signed/unsigned distance functions~\cite{park2019deepsdf,gropp2020implicit,novello2022exploring, chen2019learning, Michalkiewicz2019deep,chibane2020ndf} into neural networks. One can query any 3D points in space to obtain the corresponding attributes. This approach allows for recovering highly detailed surfaces at a lower memory cost than traditional surface representation methods such as classical signed distance field (SDF). Moreover, it is a continuous representation suitable for further mathematical analysis of the represented surface. 
A distance field stores the shortest (signed/unsigned) distance of a given point to a surface. 
The gradients of an SDF provide a normal vector of the surface, but using SDF requires well-defined "inside" and "outside" notions, which restricts representing open surfaces. An unsigned distance field (UDF)~\cite{richa2022unsigned} is devoid of this specified concern. However, the lack of a sign introduces ambiguity in surface reconstruction, such as the flipped surface shown in~\cref{fg:sampling_visual}. Consequently, the development of a novel approach is crucial to integrate the benefits of both signed and unsigned distance fields.

\subsection{Surface Reconstruction}
Similar to surface representations, surface reconstructions techniques can be categorized into traditional approaches, including \textbf{explicit} methods like Poisson surface reconstruction~\cite{Kazhdan2006poisson} and SSD~\cite{Calakli2011}. 
However, these methods heavily depend on the quality of the input data and often struggle with complex geometries or sparse and noisy inputs. Furthermore, adjusting the output resolution with traditional methods requires repeating the entire reconstruction process, which is a significant drawback. On the other hand, \textbf{implicit} surface reconstruction approaches, such as Marching Cubes~\cite{lorensen1987marching}, rely on either classical discrete voxel grids or neural networks to derive surface features~\cite{chibane2020implicit, gropp2020implicit, sitzmann2019siren, Michalkiewicz2019deep,park2019deepsdf, peng2021shape}. Recent advances in implicit reconstruction methods, combined with neural surface representation, facilitate changing the output resolution without the need to retrain the network or update the voxel grid. These methods are also more adept at handling complex geometries. Typically, the input for these methods is 3D data, such as point clouds or meshes~\cite{park2019deepsdf, huang2023neural}. Some methods also need point clouds with normals (oriented point clouds) to ensure satisfactory results~\cite{sitzmann2019siren,gropp2020implicit}. Some need ground truth meshes to provide supervision~\cite{peng2021shape,novello2022exploring}. Additionally, certain approaches incrementally use an initialized mesh for reconstruction instead of directly learning the implicit representation~\cite{peng2021shape, Hanocka_2020}. 
\subsection{Training Data Sampling}\label{subsec:sample_problem}
\inparagraph{Sparse Training Data} Learning-based methods sample data from input to train the neural implicit network.
Sampling training data is easy when meshes are given~\cite{novello2022exploring}, since it allows for infinite sampling. When only limited samples are available, such as point clouds~\cite{sitzmann2019siren, gropp2020implicit}, the commonly used random sampling does not consider the data distribution and ignores the local shape geometry characteristic, however complex geometry area may need more learning attention. This gap may lead to bad reconstruction results~\cite{chibane2020ndf, chibane2020implicit}. 
\par
\inparagraph{Biased Sampling} Another problem of random sampling is that point clouds, especially those acquired directly from the real world, are not uniformly located on the surface~\cite{yang2021geometry, novello2022exploring}. Moreover, points extracted from the iso-surface will likely be gathered near high-curvature areas~\cite {yang2021geometry}. Plus, complex surface areas need more points to represent their features. Random sampling does not consider these effects.
To avoid this issue, Yang \etal~\cite{yang2021geometry} samples on and near the iso-surface with some tolerance.
Novello \etal~\cite{novello2022exploring} proposes to sample according to principal curvatures of the surface points such that the sampled points are evenly distributed according to the curvatures. They divide points into different curvature categories according to the absolute sum of two principal curvatures. However, \emph{ground-truth meshes are required} in their computation pipeline.
\par
\inparagraph{Input Data Acquisition} The most commonly used input data to train neural implicits are point clouds or meshes. However, it is hard to get noise-free point clouds in real applications, let alone oriented point clouds. The most handy way to acquire point clouds in the real world is to use \textbf{RGB-D} sensor. Most previous papers do not discuss the input data acquisition step. Other depth-based methods focus on camera trajectory estimation~\cite{newcombe2011} or accurate depth fusion~\cite{weder2020routfusion}. 

In our work, we utilize depth as input but the fundamental differences between our work with other depth-based methods are twofold: firstly, we use depth images for geometrical feature computation that helps us to generate effective training samples. Secondly, our goal is to obtain the continuous neural implicit representation of the underlying surface captured by depth images.

To tackle the above challenges, 
we propose a method that computes an intermediate coarse voxel grid from depth images, effectively addressing the issues related to input points with normals. The voxel grid allows to locate query points in space and obtain the corresponding attributes such as SDF value by an efficient Taylor approximation, thereby resolving the sparse input problem. 
Moreover, we embed an uncertainty value into implicit surface representation to indicate the reliability of the SDF value. The uncertainty value helps to eliminate redundant areas and enables the reconstruction of open surfaces.

\section{Method}\label{sec:method}
Our goal is to train a network $f(\vt{x}, \theta): \mathbb{R}^3 \to \mathbb{R}\times [0,1]$, which predicts the SDF value together with its uncertainty, such that the reconstructed surface $\surf{S}$ lies on the level-set $\{\vt{x} | f(\vt{x}) = 0\}$. To achieve this goal, only a set of depth images $\{D_k\}$ of the object (or scene) of interest is required as input, which resembles a highly relevant application scenario in practice.\par
\begin{figure}[t]
    \centering
    \includegraphics[width=\linewidth, clip, trim=0cm 0.2cm 0cm 0cm]{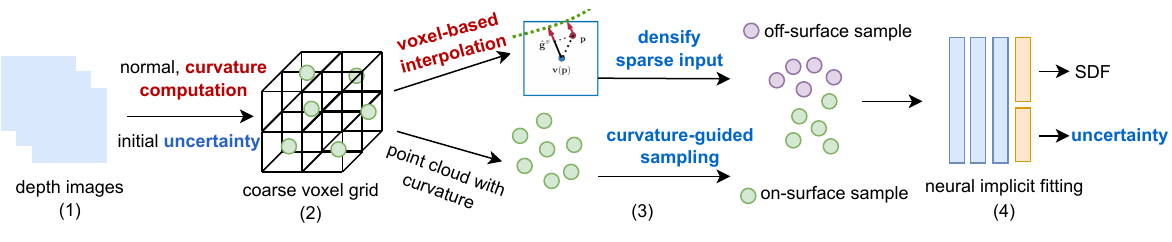}
     \vspace{-0.5cm}
    \caption{The summary of our pipeline. The \textbf{\textcolor{drawred}{red}} and  \textbf{\textcolor{drawblue}{blue}} correspond to proposed theoretical and architectural improvement. After a customized coarse voxel initialization with uncertainty $w^v$, curvature $H$, and normal $\hat{\vt{g}}^v$, we use curvature-guided sample on the extracted point cloud and using voxel-based sampling to generate more training points in the space. }
    \label{fig:pipeline}
    \vspace*{-0.6cm}
\end{figure}
A full pipeline of our method is illustrated in~\cref{fig:pipeline}. In summary, we first initialize a coarse voxel grid based on input depth images and further augment the voxel representation by integrating its corresponding curvature and uncertainty (\cref{subsec:voxelization}). After the coarse voxel grid has been prepared, we sample points from the grid to generate data. We use curvature-guided sampling for on-surface samples to overcome the unevenly distributed points problem and use the off-surface sampling method to create samples in arbitrary positions, to solve the sparse training data problem (\cref{subsec:sampling}). Finally, our network predicts uncertainty together with SDF values. We show that we can therefore extract non-watertight surfaces for open surfaces (\cref{subsec:our_method}). 
We also discuss the possibilities of incorporating our proposed techniques into existing methods and show a concrete example in the case of NeuralPull~\cite{ma2020neuralpull} (\cref{subsec:neuraPull_curv}), showcasing the flexibility of our proposed techniques.

\subsection{Coarse Voxel Grid}\label{subsec:voxelization}
We utilize a coarse voxel grid $\{\vt{v}_i\} \subset \mathbb{R}^3$, $i \in \mathcal{V}$ following the standard TSDF fusion method~\cite{newcombe2011}. 
For each voxel $\vt{v}_i \in \mathbb{R}^3$, the SDF value $\psi^v_i \in \mathbb{R}$,
the corresponding curvature $H^v_i \in \mathbb{R}$ 
and uncertainty $w^v_i \in [0,1]$ are initialized. 
(the details of uncertainty computation are described in the supplementary.)
Inspired by~\cite{sommer2022}, we also integrate the gradient of SDF $\vt{g}_i^v \in \mathbb{R}^3$  for each voxel using the computed normal of each pixel in the depth images, as the SDF gradient is the normal direction of the underlying surface.
Different from~\cite{sommer2022}, we do not use a hash map to store voxels but a regular grid, which enables off-surface sampling (we will elaborate in~\cref{subsec:sampling}). Moreover, we only require a coarse voxel grid (\eg, of size $64^3$ in our experiments) at this step, while being able to preserve details in the reconstruction, thanks to our efficient sampling strategy (\cf~\cref{subsec:sampling}).
With stored gradients in voxel elements,
the corresponding point cloud of the object contained in the voxel grid can simply be extracted in one step (without any mesh extraction) by 
\begin{equation}\label{eq:extract_point}
    \vt{x}_i = \vt{v}_i -\hat{\vt{g}}^v_i\psi^v_i, ~~~
 \text{where}~~~\hat{\vt{g}}^v_i = \frac{\vt{g}_i^v}{\|\vt{g}_i^v\|}.
\end{equation}

\inparagraph{Direct Curvature From Depth} Depth images can provide more geometric information other than normals~\cite{kurita1999, martino2014}. The mean curvature and other differential geometry features, such as the Gaussian curvature, are local geometrical properties of the surface and reveal the local topological characteristics. The mean curvature $H$ is the average of the principal curvatures, while the Gaussian curvature $K$ is their product. We propose a method that computes curvature information directly based on (discrete) depth images \emph{without (continuous) ground-truth meshes}. A depth image $D$ can be viewed as a Monge patch of a surface, i.e. $z = D(m,n), (m,n) \in \Omega \subset \mathbb{R}^2$ with pixel coordinate $(m,n)$ lies in the image domain $\Omega$~\cite{carmo1976, spivak1999comprehensive}. Hence the Monge patch $\mathcal{M}: \Omega \to \mathbb{R}^3$ is defined as 
$\mathcal{M}(m,n) = (m, n, D(m,n))$.
The two types of curvatures can be computed by
\begin{align} \label{eq::curv_depth}
&K(m,n)  = \frac{D_{mm}D_{nn} - D_{mn}^2}{(1+D_m+D_n)^2} \,, \\
&H(m,n)  = \frac{(1+D_m^2)D_{nn} - 2D_m D_n D_{mn} + (1+D_n^2)D_{mm}}{2 (1+D_m^2 + D_n^2)^{3/2}} \,, \label{eq::curv_depth2}
\end{align}
where $D_m = \frac{\partial}{\partial m} D(m,n)$ is the partial derivative of depth w.r.t. $x$-axis. Similarly $D_n = \frac{\partial}{\partial n} D(m,n)$, $D_{mn} = \frac{\partial^2}{\partial m \partial n} D(m,n)$ and other second order derivatives.
The computation is done on the fly per depth image. To our knowledge, we are the first to compute mean curvatures directly from depth images and integrate them into voxel grids for efficient sampling afterward (\cf~\cref{subsec:sampling}).
After associating a curvature to each pixel of depth images, we unproject 2D pixels into 3D space to determine its corresponding voxel element. Note that we can fuse curvatures which are computed under image coordinates, into world (voxel) coordinates due to the mean curvature $H(m,n)$ and Gaussian curvature $K(m,n)$ are invariant to changes of the parameterization on the smooth surface represented by $\mathcal{M}(m, n)$~\cite{carmo1976}. A detailed explanation is provided in the supplementary.
\cref{fg::curvature} illustrates that our estimated curvature indeed captures the local geometric properties of the surface. \par
\inparagraph{Voxel attributes initialization} For each voxel, we use a weighted averaging to iteratively update its gradient $\vt{g}^v$, curvature $H^v$, SDF $\psi^v$, and uncertainty $w^v$ information.
This process helps to reduce the error caused by noisy and sparse depth images (\cf supplementary for details).
The time required for normal and curvature computation of a $480 \times 640$ depth image, plus updating voxel attributes according to this incoming depth, is $\sim$\textbf{50ms}.
If camera poses are not available from the given depth images, we run camera pose estimation steps together with the voxelization step (see~\cite{sommer2022} and our supplementary). The whole voxelization step plus camera pose estimation (when ground truth camera poses are absent) takes $\sim$\textbf{120ms} per depth image.

\begin{figure}[t]
   \begin{minipage}{0.36\linewidth}
   \subcaptionbox{\label{fg::curvature}}{
       \centering
        \begin{tabular}{m{0.5\linewidth} m{0.5\linewidth}}
            \includegraphics*[width=0.9\linewidth]{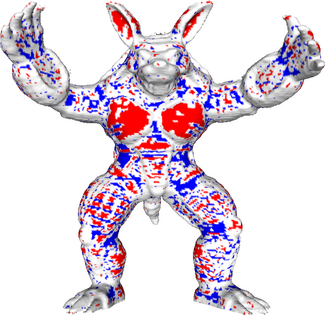} &
            \includegraphics*[width=0.9\linewidth]{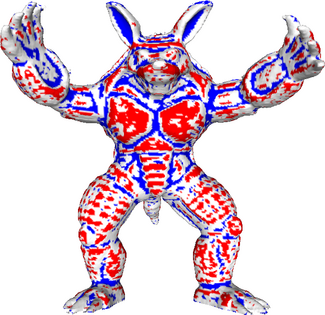}\\
            \centering \scriptsize{$K$} & \centering \scriptsize{$H$}
        \end{tabular}
        }
   \end{minipage}%
   \begin{minipage}{0.30\linewidth}
   \subcaptionbox{\label{fg::sample_points}}{
   \begin{tabular}{m{0.5\linewidth} m{0.5\linewidth}}
            \includegraphics*[width=0.85\linewidth]{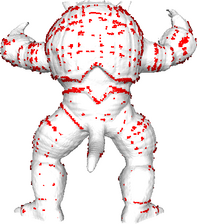} &
            \includegraphics*[width=0.85\linewidth]{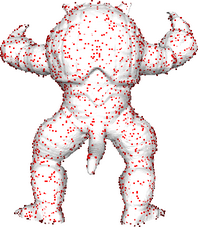}
        \end{tabular}
}
   \end{minipage}
   \begin{minipage}{0.33\linewidth}
   \centering
    \subcaptionbox{\label{fg:taylor}}{
        \includegraphics[width=0.9\linewidth]{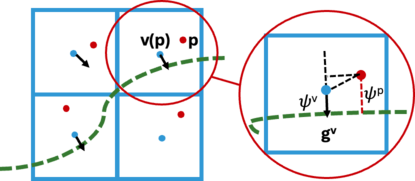}}
   \end{minipage}
   \caption{(a) The visualization of Gaussian curvatures and mean curvatures of each point. The red color indicates a high curvature area, and the blue color indicates a low curvature area. A positive mean curvature ($H > 0$) signifies a convex surface, and a negative mean curvature ($H < 0$) indicates a concave surface. Positive Gaussian curvature ($K > 0$) indicates that the surface is locally like a dome, and negative Gaussian curvature ($K < 0$) indicates that the surface is locally saddle-shaped. (b) Points gathering on high curvature effect (left) and our sampling results after considering points mean curvature (right). (c) Illustration of~\cref{eq:sdf_sample}. For each query point that falls in voxel $\vt{v}$, the SDF value is the SDF value of the voxel $\psi^v$ plus the projected distance of the voxel center to the query point to the gradient direction. }
   \vspace{-0.7cm}
\end{figure}
\subsection{Efficient Sampling Strategy}\label{subsec:sampling}
To fit a neural implicit function, pairs of 3D points and the corresponding SDF values are often needed as the supervision~\cite{park2019deepsdf, novello2022exploring, chibane2020implicit}. Many previous works only sample points in a point cloud~\cite{sitzmann2019siren, gropp2020implicit} or need ground truth mesh to compute the signed distance when sampling in the space~\cite{novello2022exploring}. 
All these methods produce unsatisfactory results when input data are sparse and ground truth mesh is not available. 
In this section, we introduce an interpolation strategy that deals with the sparse input and deploys the estimated gradient and curvature stored in the coarse voxel grid to generate both on- and off-surface samples, which is crucial for effective neural fitting, as shown in our extensive experiments.\par
\inparagraph{On-Surface Sampling} To sample points on the surface, one naive approach is to uniformly (random) sample on surface points utilized using~\cref{eq:extract_point}. However, the surface points extracted by this step are often concentrated at the high curvature area (see \cref{fg::sample_points} left)~\cite{yang2021geometry} and this uneven distribution will be inherited by the uniform sampling during training. To avoid this uneven sampling problem, we divide sampled points into low, median, and high curvature regions based on the mean curvature, similarly as in \cite{novello2022exploring}. Note that the surface point extracted by~\cref{eq:extract_point} inherit the integrated mean curvature $H$. Moreover, the Gaussian curvature can be employed as well and we prove that these two metrics are equivalent (see \cref{fg::curvature} and supplementary). 
Every epoch, $m$ points are sampled in each of three curvature regions defined by the thresholds $\underbar{H}$ and $\bar{H}$, which are chosen to be $0.3$ and $0.7$ percentile of the entire curvature range. \par
\inparagraph{Off-Surface Sampling}  
To sample point $\vt{p}$ random in space and directly use it for neural implicit fitting, its SDF value $\psi^p$ is required. This is easy to achieve when the input is mesh since one can easily compute the point-to-mesh distance with a sign. However, when only a discrete structure is available, sampling off the surface is not straightforward. Here we introduce an off-surface sampling strategy. 
With the coarse voxel grid $\{\vt{v}_i\}$ estimated in~\cref{subsec:voxelization}, we can randomly sample points $\vt{p} \in \Gamma$, where $\Gamma \subset \mathbb{R}^3$ is the space defined by the voxel grid. 
The voxel element corresponding to the sampled point $\vt{p}$ can be localized by a simple operation $\vt{v}(\vt{p}) = \round{\vt{p}/v_s}$, where $\round{\cdot}$ is the rounding operator, $\vt{v}(\vt{p})$ and $v_s$ are the coordinate of the center and the size of the voxel cube respectively. Instead of inheriting the SDF value of the corresponding voxel $\psi^v$, we approximate it by the first-order Taylor expansion. Due to the stored gradients $\hat{\vt{g}}^v$ in the voxel grid, a randomly sampled point its SDF $\psi^p$ this can be computed simply by (see~\cref{fg:taylor} for an illustration):
 \begin{equation}\label{eq:sdf_sample}
\psi^p = \psi^v + \dotp{\hat{\vt{g}}^v, \vt{p} - \vt{v}(\vt{p})} \,,
\end{equation}
The uncertainty of $\vt{p}$ is interpolated using 
\begin{equation}\label{eq:weight_sample}
    w^p = \frac{v_s - \psi^p}{v_s} w^v \,,
\end{equation}
where $w^v$ is the voxel uncertainty
estimated in \cref{subsec:voxelization}. 
Thanks to the stored gradient, our method does not need any trilinear interpolation or additional nearest neighbor search. It also enables \emph{continuous} sampling in a \emph{discrete} voxel grid, and at the same time retains as much information as possible.
~\cref{eq:weight_sample} sets the maximal possible uncertainty $w^v$ to points on the surface (i.e. when $\psi^p = 0$) and reduces uncertainty when points are moving away from the surface since the accuracy of Taylor approximation reduces for distant samples away from the surface. Meanwhile, the point $\vt{p}$ inherits the normal of the voxel $\hat{\vt{g}}^p = \hat{\vt{g}}^v$.
\subsection{Uncertainty-Aware Neural Implicit Function}\label{subsec:our_method}
\inparagraph{Our Loss} 
To recover the neural implicit function $f:\mathbb{R}^3 \to (\psi, w) \subset \mathbb{R}\times[0,1]$, such that the surface lies on the level-set $\{\vt{x} | f(\vt{x}) \in 0 \times (\tau,1] \}$,
where $f^{-1}(\{(\psi, w)\in \mathbb{R}\times[0,1] \: | \: \psi = 0\})$ 
and $w$ is the uncertainty of the predicted signed distance value $\psi$. $\tau$ is the uncertainty threshold.
Given the points sampled in the voxel grid from the previous subsection and their associated SDF value and uncertainty (~\cref{eq:sdf_sample} and~\cref{eq:weight_sample}), we have all the ingredients to train our neural implicit function, which predicts both the SDF value of the query points and its corresponding uncertainty. Finally, our total loss incorporating the geometric and the normal constraints reads
\begin{align}
    \mathit{l}_\mathcal{X}(\theta) &= \frac{1}{|\Gamma^+|}\int_{\Gamma^+} |\psi - \psi^p|d\Gamma \,, \label{eq:geo_loss} \\
    \mathit{l}_{\mathcal{W}}(\theta) &= \frac{1}{|\Gamma|}\int_{\Gamma} |w - w^p|d\Gamma \,, \label{eq:weight} \\
    \mathit{l}_{\mathcal{N}} (\theta) &= \frac{1}{|\Gamma^+|} \int_{\Gamma^+} (1- \dotp{\frac{\nabla_{\psi} f(\vt{p}, \theta)}{\norm{\nabla_{\psi} f(\vt{p}, \theta)}}, \hat{\vt{g}}^p})d\Gamma \,, \label{eq:normal_ours} \\
    \mathit{l}_{\mathcal{E}}(\theta) &= \frac{1}{|\Gamma|}\int_{\Gamma} |\norm{\nabla_\psi f(\vt{p},\theta)}^2 - 1| d\Gamma \,. \label{eq:eikonal}
\end{align}
where $\Gamma^+$ indicates the area with the sampled uncertainty $w^p>0$ and $\theta$ is the network parameter. 
\begin{equation}\label{eq:loss}
    \mathit{l}(\theta, \theta_r) = \mathit{l}_\mathcal{X}(\theta) + \tau_n \mathit{l}_{\mathcal{N}} (\theta) + \tau_w \mathit{l}_{\mathcal{W}}(\theta_r) + \tau_e \mathit{l}_{\mathcal{E}}(\theta) \,.
\end{equation}
The terms $\mathit{l}_\mathcal{X}$ and $\mathit{l}_{\mathcal{W}}$ 
penalize the SDF value and uncertainty differences between the network prediction and the pseudo ground truth interpolated by~\cref{eq:sdf_sample} and~\cref{eq:weight_sample}. 
The normal term $\mathit{l}_{\mathcal{N}}$ term evaluates the cosine similarity of surface normal and implicit function gradient.
The Eikonal term $\mathit{l}_{\mathcal{E}}$ regularizes the underlying function to represent a valid signed distance field.

\setlength{\columnsep}{7pt}%
\setlength{\intextsep}{0pt}%
\begin{wrapfigure}{r}{0.44\linewidth}
  \begin{minipage}[b]{0.30\linewidth}
    \centering
   \includegraphics[width=0.8\linewidth]{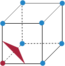}
  \end{minipage}%
  \begin{minipage}[b]{0.30\linewidth}
    \centering
    \includegraphics[width=0.8\linewidth]{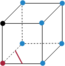}
  \end{minipage}%
  \begin{minipage}[b]{0.30\linewidth}
    \centering
    \includegraphics[width=0.8\linewidth]{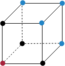}
  \end{minipage}
  \caption{surface extraction with uncertainty. Black vertices represent zero uncertainty points. Red and blue vertex mean points with negative and positive SDF values, respectively.}
  \label{fg:cubes}
\end{wrapfigure}
\inparagraph{Open Surface Reconstruction} While signed distance fields are flexible in representing objects with different topologies, they require the underlying object to be water-tight. On the other hand, uncertainty can serve as an open area indicator. When extracting surface using the Marching cubes algorithm~\cite{lorensen1987marching}, we skip the area where the uncertainty is under a threshold $w_c$ (\eg $w_c$=0). 
As shown in~\cref{fg:cubes}, 
a single zero uncertainty vertex leads to a line (instead of a triangle), while two of this kind lead to a point~\cite{botsch2010polygon, farin2002curves}. Therefore, we can naturally reconstruct open surfaces, thanks to our uncertainty estimation. 

\subsection{Incorporating with Other Methods}\label{subsec:neuraPull_curv}

Our curvature-guided sampling can be seamlessly incorporated into popular implicit surface reconstruction methods such as~\cite{gropp2020implicit,sitzmann2019siren}. These methods take point clouds as input. The points with curvature can be extracted using Eq.~\eqref{eq:extract_point} is just $\psi^p =0$ situation.
Additionally, our interpolating method simplifies the method that needs nearest neighbor search, such as NeuralPull~\cite{ma2020neuralpull}. NeuralPull proposes to train a network to learn to pull a query point to the closest iso-surface. During training, the method samples random points $\vt{p}$ and uses nearest neighbor search to find the closest surface point $\vt{x}$. Then, a network $f$ is trained to minimize the loss.
\begin{equation} \label{eq:neural_pull_loss}
    \mathit{l}_\mathcal{X}(\theta) = \frac{1}{|\Gamma|}\int_{\Gamma} \norm{ \vt{x} - \vt{p} + \frac{\nabla f(\vt{p})}{\norm{\nabla f(\vt{p})}}f(\vt{p}) } d \Gamma \,,
\end{equation}
with $\tau_n=0$, $\tau_e=0$.
Our interpolating method (\cref{subsec:sampling}) eliminates the nearest neighbor search required in the original NeuralPull. As for a random point $\vt{p}$, after finding the corresponding voxel $\vt{v}(\vt{p})$, the closest surface point $\vt{x}$ can be easily located using~\eqref{eq:extract_point}. In~\cref{sec:evaluation}, we show that it effectively reduces the noise during training and leads to better reconstruction quality. The modified NeuralPull outperformed the original implementation, especially under sparse inputs.

\section{Evaluation}\label{sec:evaluation}

To demonstrate that our proposed method improves the robustness and accuracy of the neural implicit fitting, we validate our method extensively on both \textbf{synthetic} and \textbf{real-world} datasets, including \textbf{objects} and \textbf{scene} scenarios. 
For synthetic datasets with ground-truth mesh, we rendered noise-free depth images and camera poses to compute the coarse voxel grid $\{\vt{v}_i\}$. 
To test different quality inputs, we compare two different voxel resolutions, $64^3$ (\textbf{coarse}) and $256^3$ (\textbf{dense}). The coarse voxel grid needs $25$ MB while the dense one takes $1.5$ GB of memory. 
The real-world datasets contain RGB-D sequences with noisy, sparse depth images and noise camera poses. Using this setting, we show that even with the intermediate coarse voxel grid, our method can still reconstruct faithfully. 
We use an 8-layer multi-layer perceptron (MLP) with ReLU activation. 
Each layer has $256$ nodes, and the last layer has $2$ output heads for the SDF value and its uncertainty respectively.
Note that due to the space constraint, we focus on showing the comparison methods. Ablation study of our loss function together with more comparison details please refer to our supplementary materials. \par
\inparagraph{Evaluation Metrics} Quantitative evaluation is performed on four synthetic datasets, where the 3D ground truth is available. 
We compute the Chamfer distance (CD)~\cite{de2000computational} and Hausdorff distance (HD)~\cite{hausdorff2005set} of the reconstruction \wrt the ground-truth.




\subsection{Comparison with Depth-Based Methods}

\input{figures_tex/discrete_analysis}
\setlength{\intextsep}{0pt}%

 \begin{table}[t]
     \centering
        \scriptsize
     \begin{tabular}{|p{2cm}|cccc|cccc|}
     \toprule
          & \multicolumn{4}{c|}{\textbf{CD}$\downarrow$($\times 10^2$)} & \multicolumn{4}{c|}{\textbf{HD}$\downarrow$} \\
          \cline{2-9}
          \textbf{Method} & Bunny & Armadillo & Dragon & Buddha &  Bunny & Armadillo & Dragon & Buddha \\
          \midrule
          RoutedFusion & 0.267 & 0.272 & 0.578 & 0.356 & 0.116 & 0.066 & 0.163 &  0.086 \\
          TSDF & 0.073 & 0.080 & 0.220 & 0.287 & \textbf{0.014} & 0.008 & 0.038 & 0.018\\
          gradient-SDF & 0.111 & 0.210 & 0.165 & \textbf{0.118} & 0.028 & 0.027 & \textbf{0.028} & 0.024\\
          Ours &  \textbf{0.067} & \textbf{0.037} &\textbf{0.157} & 0.125 &0.016 &\textbf{0.006} & 0.040 & \textbf{0.017} \\
          \bottomrule
     \end{tabular}
     \caption{ Quantitative comparison with depth-based methods. The best accuracy is bold-faced for each dataset. 
     }
     \label{tab:depth_metric}
     \vspace*{-1.0cm}
 \end{table}
We first compare with both classical and learning-based depth-based methods. TSDF fusion~\cite{newcombe2011} is one of the standard works in depth fusion. Gradient-SDF~\cite{sommer2022} and RoutedFusion~\cite{weder2020routfusion} are two recent representative classical and learning-based methods respectively.
For TSDF fusion, we employ a voxel grid of $128^3$. Despite our initial input having only \emph{half the resolution} of the TSDF, we achieve superior outcomes. Gradient-SDF employs a hash map to store individual voxel data, which constructs voxels solely around the projected depth points. This method doesn't specify a fixed voxel resolution; instead, it directly controls the voxel size to control the resolution. Hash map plus gradient stored in voxels helps to reduce the memory and computational time, but the discretely stored voxels make the method unreliable for sparse and noisy depth.
RoutedFusion~\cite{weder2020routfusion} leverages two networks—Routing and Fusion—utilizing both \emph{ground truth camera poses} and \emph{ground truth or previously derived meshes from TSDF} to facilitate their guided fusion. Hence we input noise-free depth images and ground truth camera poses for the experiments of RoutedFusion. Despite its high requirements for the inputs, RoutedFusion creates dense and high-resolution meshes however with a lot of artifacts around it. One reason could be that since the method trains one network for fusing different objects from the same dataset (similar shapes, like ShapeNet~\cite{chang2015shapenet}), however, in our synthetic datasets the object sizes have very large varies. The method fails to generalize. 
The quantitative results can be found in~\cref{tab:depth_metric} and the visualization results are shown in~\cref{fig:discrete_analysis} (more analysis and results on other datasets please refer to supplementary). We also demonstrate the difference between TSDF (discrete SDF) and Ours (continuous SDF) by visualizing the resulting signed distance fields 

\subsection{Efficient Sampling Comparison}
\begin{figure}[t]
    \centering
   \begin{tabular}{m{0.1\linewidth}m{0.1\linewidth}m{0.1\linewidth}m{0.1\linewidth}m{0.1\linewidth}m{0.1\linewidth}m{0.1\linewidth}m{0.1\linewidth}m{0.1\linewidth}}
    \includegraphics*[width=\linewidth]{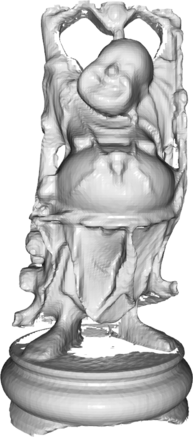} &
    \includegraphics*[width=\linewidth]{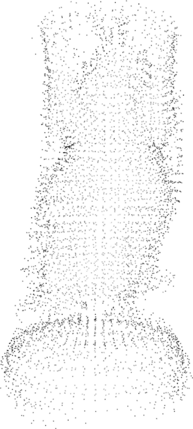} &
    \includegraphics*[width=\linewidth]{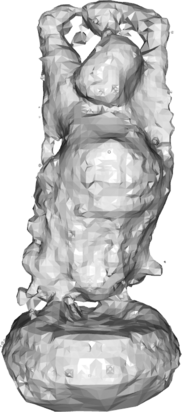}&
    \includegraphics*[width=\linewidth]{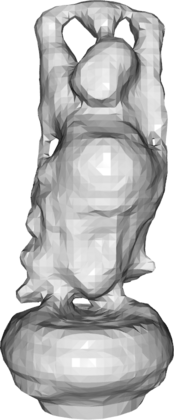} &
    \includegraphics*[width=\linewidth]{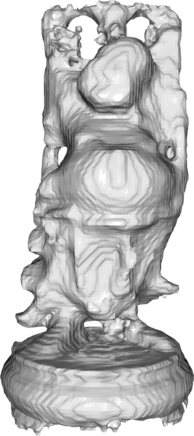} & 
    \includegraphics*[width=\linewidth]{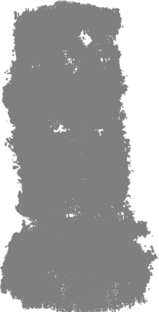} &
    \includegraphics*[width=\linewidth]{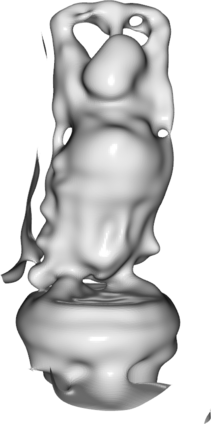} &
    \includegraphics*[width=\linewidth]{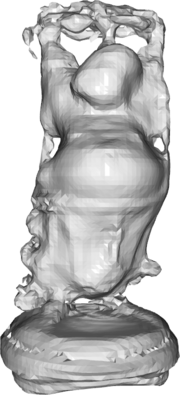} &
    \includegraphics*[width=\linewidth]{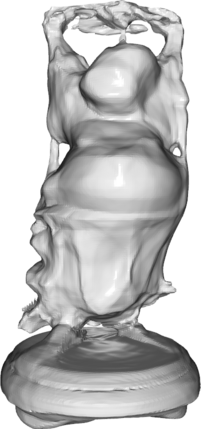} \\
    \includegraphics*[width=\linewidth]{images/happy_budda/64/init_mesh_crop.png} &
    \includegraphics*[width=\linewidth]{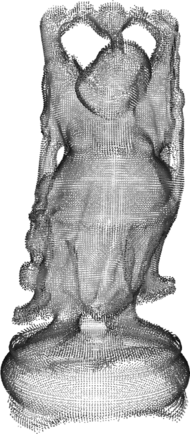} &
    \includegraphics*[width=\linewidth]{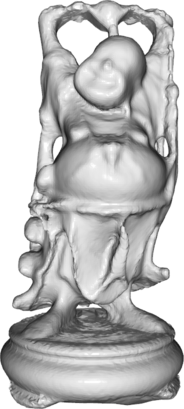}&
    \includegraphics*[width=\linewidth]{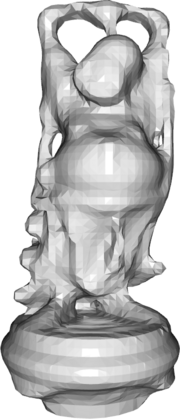} &
    \includegraphics*[width=\linewidth]{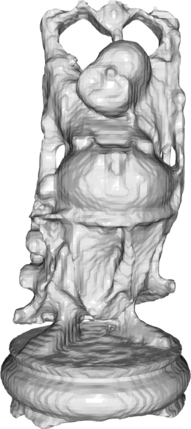} & 
    \includegraphics*[width=\linewidth]{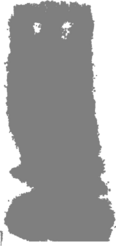} &
    \includegraphics*[width=\linewidth]{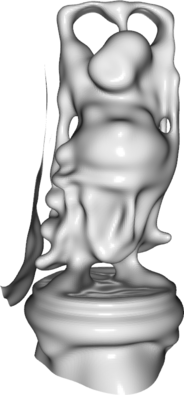} &
    \includegraphics*[width=\linewidth]{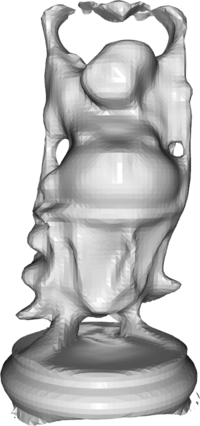} &
    \includegraphics*[width=\linewidth]{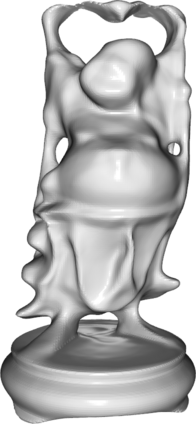} \\
 \centering \scriptsize{GT mesh} & \centering Input & \centering SSD & \centering Poisson & \centering IF-NET & \centering NDF & \centering SIREN & \centering IGR & {\centering Ours}
   \end{tabular}
   \caption{Comparison results with IGR~\cite{gropp2020implicit}, SIREN~\cite{sitzmann2019siren} and IF-NET~\cite{chibane2020implicit} with two different density input on synthetic datasets. Sparse input \emph{happy\_buddha}~\cite{bunny} has $\sim 5k$ points, and dense one has $\sim 96k$ points.
   }\label{fg:sampling_visual}
   \vspace*{-0.4cm}
\end{figure}

\inparagraph{Off-Surface Point Sampling} To evaluate the impact of off-surface sampling on the training  of neural implicits and subsequent surface reconstruction—particularly in scenarios of sparse input—we conducted comparisons with traditional methods such as the smooth signed distance (SSD) method~\cite{Calakli2011} and Poisson surface reconstruction~\cite{Kazhdan2006poisson}, both of which generate polygon meshes from point cloud inputs. Additionally, for comparison within the learning-based neural implicit methods, we selected IGR~\cite{gropp2020implicit}, IF-NET~\cite{chibane2020implicit}, NDF~\cite{chibane2020ndf}, and SIREN~\cite{sitzmann2019siren}. Unlike NDF, which predicts the unsigned distance value (UDF) and produces a denser point cloud, the other methods estimate the signed distance function (SDF) values, subsequently utilizing MarchingCubes to derive meshes at specified resolutions.
For the Poisson, SSD, IGR, NDF, and SIREN methods, inputs include both sparse and dense point clouds, derived from voxels using equation~\cref{eq:extract_point} to maintain consistent surface point density with normals. Meanwhile, IF-NET employs meshes generated from initialized voxels through MarchingCubes as its input.
~\cref{fg:sampling_visual} shows the visual comparison and the~\cref{tab:sample_error} shows the quantitative comparison. 
All methods produce satisfactory results when the input points are dense. 
Our method still gives satisfactory results when the input is sparse. \par

\begin{figure}[t]
    \centering
        \begin{tabular}{m{0.01\linewidth}m{0.12\linewidth}m{0.12\linewidth}m{0.12\linewidth}m{0.01\linewidth}m{0.14\linewidth}m{0.14\linewidth}m{0.14\linewidth}}
        \rotatebox{90}{\scriptsize{random}} &
        \includegraphics*[width=0.9\linewidth]{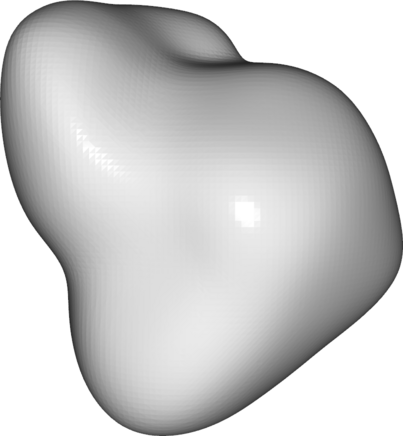} &
        \includegraphics*[width=0.9\linewidth]{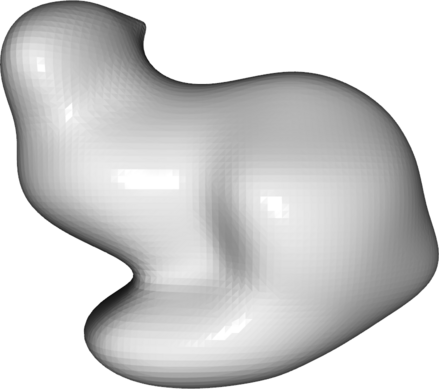} &
        \includegraphics*[width=0.9\linewidth]{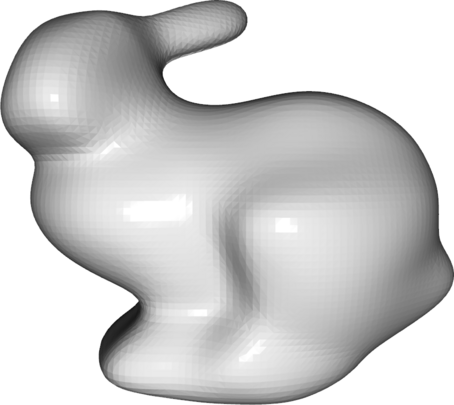} &
        \rotatebox{90}{\scriptsize{random}} &
        \includegraphics*[width=0.9\linewidth]{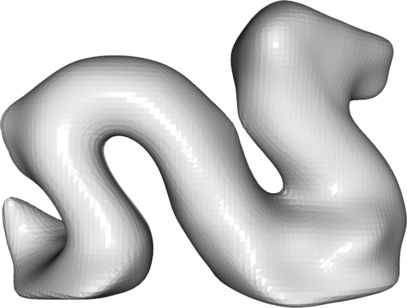}&
        \includegraphics*[width=0.9\linewidth]{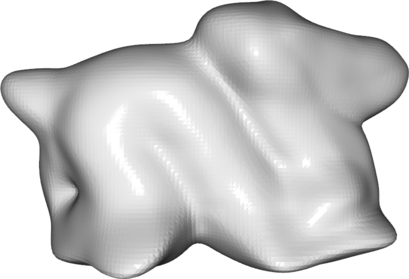}&
        \includegraphics*[width=0.9\linewidth]{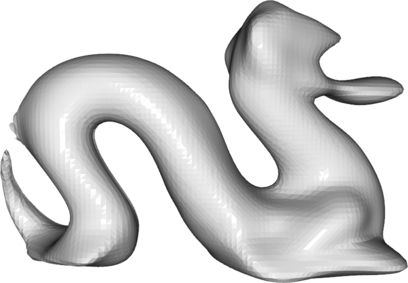} \\
         \rotatebox{90}{\scriptsize{with curv}} & 
        \includegraphics*[width=0.9\linewidth]{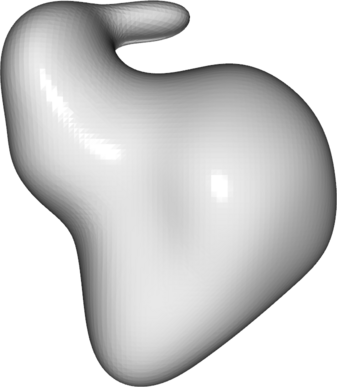} &
        \includegraphics*[width=0.9\linewidth]{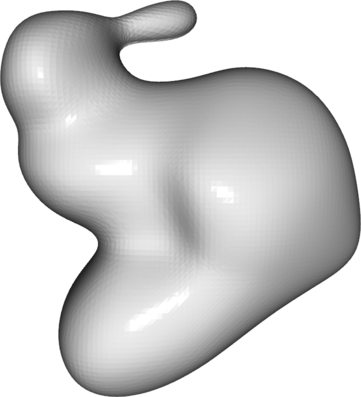} &
        \includegraphics*[width=0.9\linewidth]{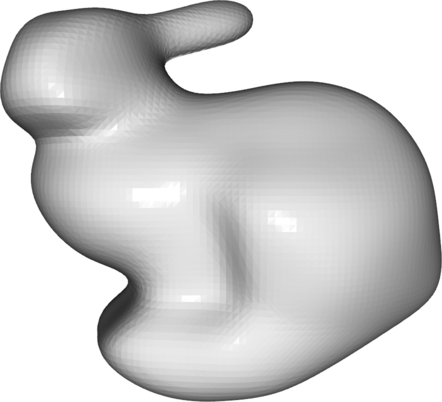} &
        \vspace*{0.1cm}\rotatebox{90}{\scriptsize{with curv}} & 
        \includegraphics*[width=0.9\linewidth]{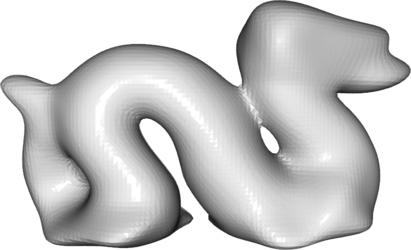}&
        \includegraphics*[width=0.9\linewidth]{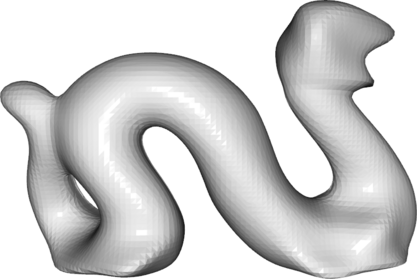}&
        \includegraphics*[width=0.9\linewidth]{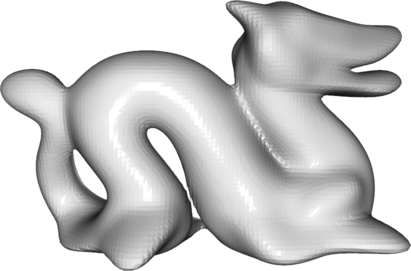} \\
        & 30 epochs & 50 epochs & 70 epochs & & 70 epochs & 90 epochs & 120 epochs
       \end{tabular}
       \caption{We extract surface during training to compare the curvature-guided sampling and random sampling.} \label{fg:curvature_epochs}
       \vspace*{-0.5cm}
\end{figure}

\begin{table}[t]
    \centering
    \scriptsize
    \begin{tabular}{|p{0.8cm}|p{1.0cm}c|cccccccc|}\toprule
    \multirow{2}{*}{\textbf{{\tiny{Metric}}}} &\multicolumn{2}{c|}{\multirow{2}{*}{\textbf{Dataset}}} & \multicolumn{8}{c|}{\textbf{Method}}\\
    \cline{4-11}
    & & & SSD & Poisson & IF-NET & SIREN & IGR & IGR (curv) & Ours & Ours (curv)\\
    \midrule
    \multirow{8}{*}{\begin{minipage}{1cm}\textbf{CD}$\downarrow$\\($\times 10^2$)\end{minipage}} & \multirow{2}{*}{Bunny} & \multicolumn{1}{|c|}{sparse} & 0.204 & 0.278 & 0.303 & 8.745 & 0.481 & 0.447 & 0.068 & \textbf{0.067} \\
    & & \multicolumn{1}{|c|}{dense} &0.224 & 0.242 & 0.315 & 7.582 & 0.501 & 0.491 & 0.073 & \textbf{0.068}  \\
    & \multirow{2}{*}{\tiny{Armadillo}} & \multicolumn{1}{|c|}{sparse} & 0.038 & 0.031 & \textbf{0.025} & 2.396 & 0.060 & 0.034 & 0.039 & 0.037  \\
    & & \multicolumn{1}{|c|}{dense} & \textbf{0.024} & 0.025 & 0.026 & 2.331 & 0.035 & 0.034 & 0.031 & 0.030 \\
    & \multirow{2}{*}{Dragon} & \multicolumn{1}{|c|}{sparse} & 0.210 & 0.206 & 0.158 & 2.710 & 0.209 & 0.198 & 0.179 & \textbf{0.157} \\
    & & \multicolumn{1}{|c|}{dense} & 0.151 & 0.150 & 0.181 & 2.784 & 0.162 & 0.154 & 0.130 & \textbf{0.126} \\
    & \multirow{2}{*}{\begin{minipage}{1cm} Happy \\ Buddha \end{minipage}} & \multicolumn{1}{|c|}{sparse} & 0.180 & 0.240 & 0.258 & 2.717 & 0.307 & 0.259 & 0.135 & \textbf{0.125}  \\
    & & \multicolumn{1}{|c|}{dense} & 0.210 & \textbf{0.202} & 0.258  &  2.720 & 0.248 & 0.240 & 0.214 & 0.267 \\
    \midrule
    \multirow{8}{*}{\textbf{HD}$\downarrow$} & \multirow{2}{*}{Bunny} & \multicolumn{1}{|c|}{sparse} & 0.057 & 0.074 & 0.112 &  0.817 & 0.149  & 0.142 & 0.026 & \textbf{0.016}  \\
    & & \multicolumn{1}{|c|}{dense} & 0.071 & 0.065 & 0.107 &  0.816 & 0.153 & 0.155 & 0.020 & \textbf{0.012}  \\
    & \multirow{2}{*}{\tiny{Armadillo}} & \multicolumn{1}{|c|}{sparse} & 0.014 & 0.006 &  0.007 & 0.357 & 0.018 & 0.008 & 0.015 & \textbf{0.006} \\
    & & \multicolumn{1}{|c|}{dense} & 0.005 & 0.006 & \textbf{0.004} & 0.301 & 0.014 & 0.013 & 0.007 & 0.007 \\
    & \multirow{2}{*}{Dragon} & \multicolumn{1}{|c|}{sparse} & 0.043 & 0.050 & 0.037 & 0.340 & 0.037 & 0.039 & \textbf{0.030} & 0.040 \\
    & & \multicolumn{1}{|c|}{dense} & 0.040 & 0.039 & 0.036 & 0.317 & 0.056 & 0.045 & 0.030 & \textbf{0.025} \\
    & \multirow{2}{*}{\begin{minipage}{1cm} Happy \\ Buddha \end{minipage}} & \multicolumn{1}{|c|}{sparse} & 0.048 & 0.061 & 0.059 & 0.338 &  0.092 & 0.081 & 0.023 & \textbf{0.017}  \\
    & & \multicolumn{1}{|c|}{dense} & 0.075 & 0.057 & 0.058 & 0.332 & 0.059 & 0.057 & 0.054 & \textbf{0.044} \\
    \bottomrule
    \end{tabular}
    \caption{Quantitative evaluation of comparison methods: We compare our off-surface sampling method (second last column) and plus on-surface curvature-guided sampling method (last column) with other comparison methods. Additionally, we show our curvature-guided on-surface sampling method incorporated with IGR, compared with the original IGR, see column IGR(curv) and IGR.}
    \label{tab:sample_error}
\vspace*{-0.8cm}
\end{table}
\inparagraph{On-Surface Point Sampling}
We show that curvature-guided sampling helps in two aspects. First, it stabilizes the learning procedure, leading to a faster convergence of the minimum solution. 
We show this visually by rendering reconstructed meshes during early training epochs to see the learning efficiency of the different sampling methods (see~\cref{fg:curvature_epochs}). We then compute these CD and HD to show that curvature-guided sampling has a smoother error curve. In~\cref{fg:curvature_curv}, the solid lines and dash lines are CD errors of curvature-guided sampling and random sampling, respectively. We normalized the CD errors by dividing the maximum error within each dataset to draw all lines in one figure. The curvature-guided sampling lines have a smoother trend and reach a lower error faster.
Second, we show that curvature-guided sampling also increases the accuracy of reconstructed meshes. We test our method (with enabled off-surface sampling) with and without curvature-guided sampling (\cref{tab:sample_error}~ Ours(curv) and Ours) for a comparison. Both comparison pairs show that considering curvature information during training improves the results. 

\subsection{Improving Other Methods with Our Sampling Strategy}
\begin{figure}[t]
    \centering
   \begin{tabular}{m{0.11\linewidth}m{0.11\linewidth}m{0.11\linewidth}m{0.11\linewidth}m{0.11\linewidth}m{0.11\linewidth}m{0.11\linewidth}m{0.11\linewidth}}
    \includegraphics*[width=\linewidth]{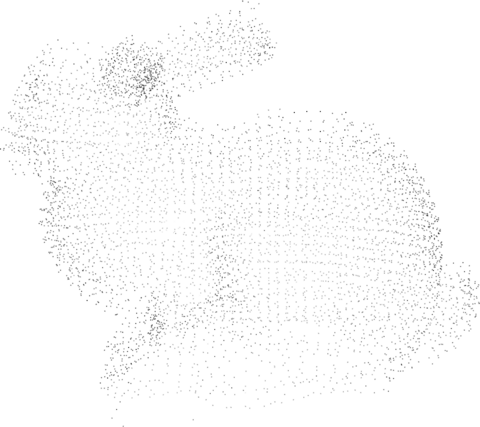} &
    \includegraphics*[width=\linewidth]{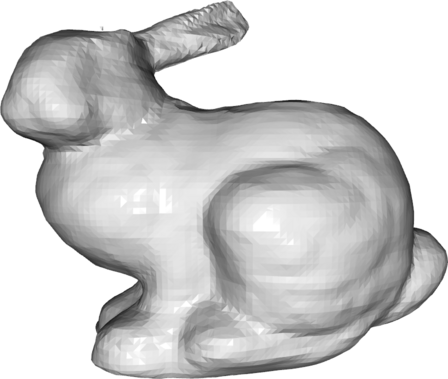} &
    \includegraphics*[width=\linewidth]{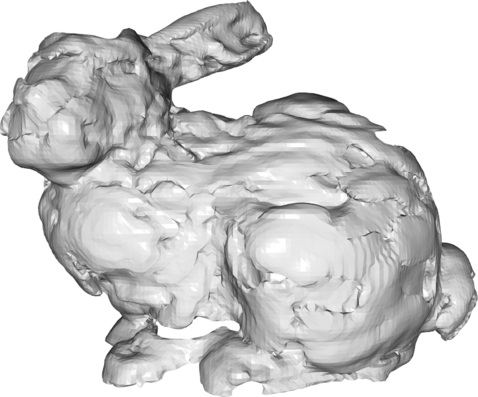} &
    \includegraphics*[width=\linewidth]{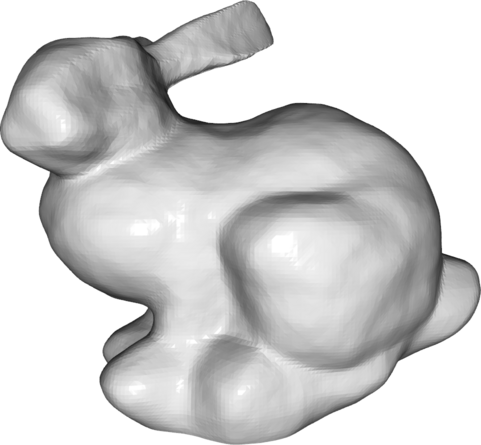} &
    \includegraphics*[width=\linewidth]{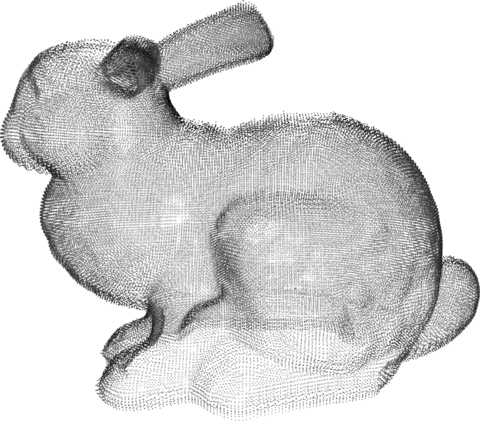} &
    \includegraphics*[width=\linewidth]{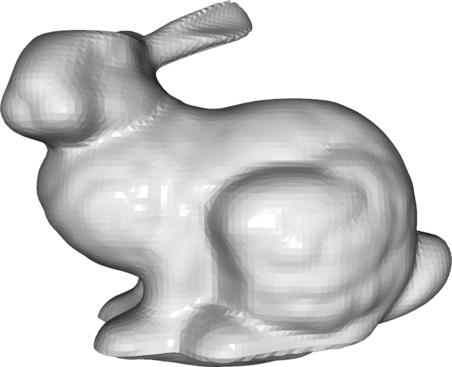} &
    \includegraphics*[width=\linewidth]{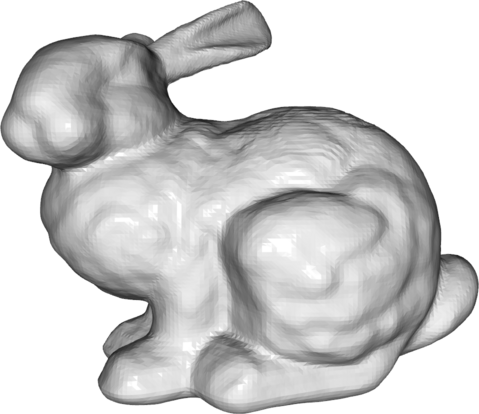} &
    \includegraphics*[width=\linewidth]{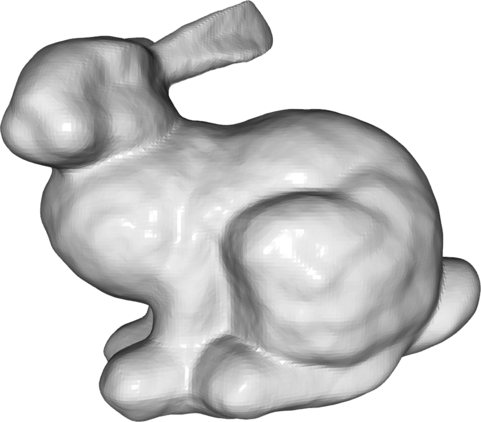} \\
    \includegraphics*[width=\linewidth]{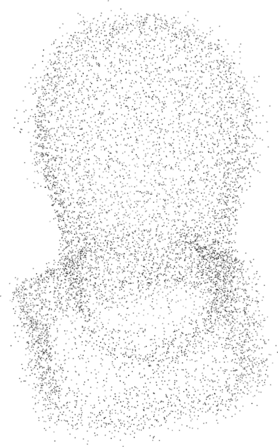} &
    \includegraphics*[width=\linewidth]{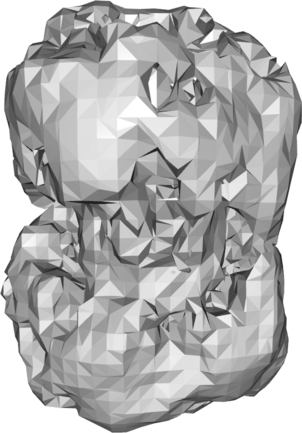} &
    \includegraphics*[width=\linewidth, angle=180, origin=c]{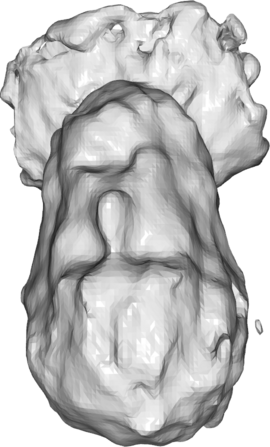} &
    \includegraphics*[width=\linewidth]{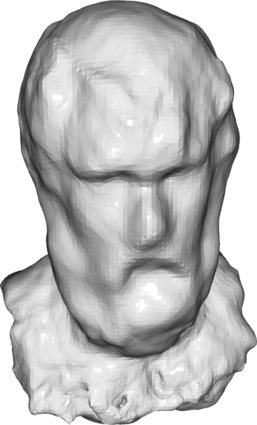} &
    \includegraphics*[width=\linewidth]{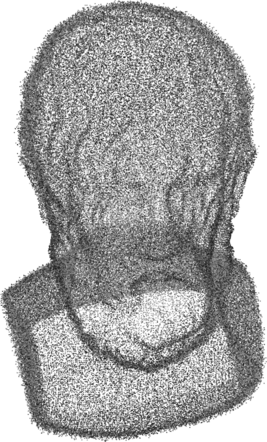} &
    \includegraphics*[width=\linewidth]{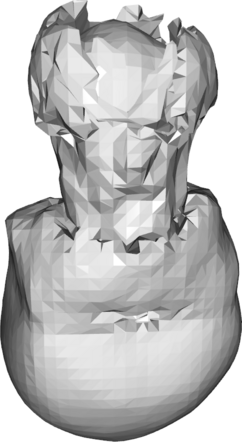} &
    \includegraphics*[width=\linewidth,angle=180, origin=c]{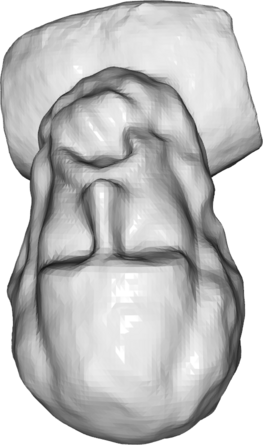} &
    \includegraphics*[width=\linewidth]{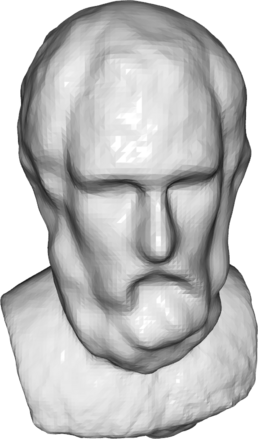} \\
    \includegraphics*[width=\linewidth]{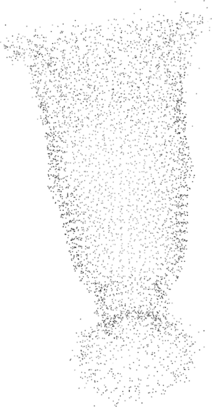}& 
    \includegraphics*[width=\linewidth]{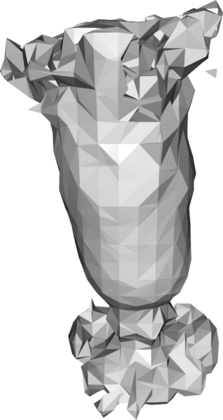} &
    \includegraphics*[width=\linewidth]{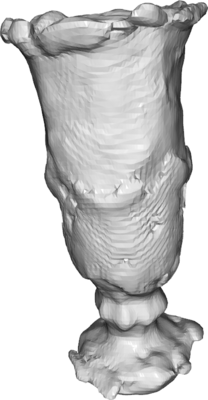} &
    \includegraphics*[width=\linewidth]{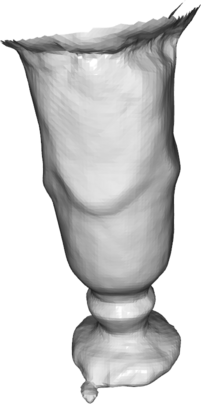} &
    \includegraphics*[width=\linewidth]{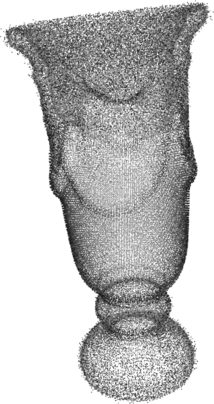} &
    \includegraphics*[width=\linewidth]{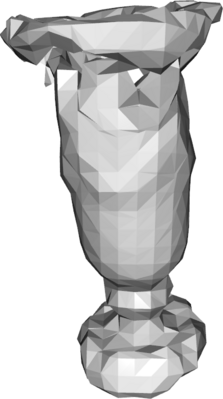} &
    \includegraphics*[width=\linewidth]{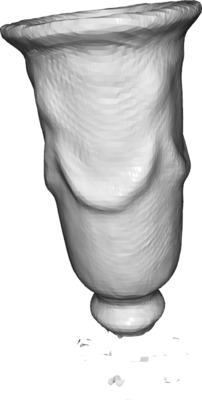} &
    \includegraphics*[width=\linewidth]{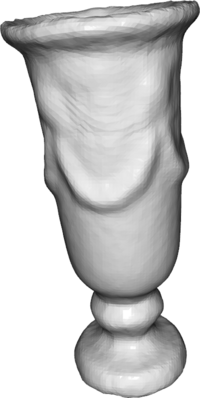} \\
    \centering \scriptsize{Sparse Input} & \centering IGR & \centering NeuralPull & \centering  Ours & \centering  \scriptsize{Dense Input} & \centering  IGR & \centering NeuralPull & \centering  Ours
   \end{tabular}
   \caption{Comparison results with NeuralPull~\cite{ma2020neuralpull} and IGR~\cite{gropp2020implicit}. Sparse input on synthetic dataset \emph{bunny} has $\sim 5k$ points and dense one with $\sim 100k$ points. Two real-world datasets \emph{sokrates} have $\sim 9k$ points and $150k$ points in sparse and dense situations, \emph{vase} has $\sim 4k$ and $\sim 81k$ respectively.}\label{fg:neural_pull}
   \vspace{-0.8cm}
\end{figure}

\inparagraph{On-Surface Point Sampling} We now show that our on-surface curvature-guided sampling can be incorporated seamlessly with previous neural fitting methods and improve accuracy. We change the sampling method in IGR~\cite{gropp2020implicit} to curvature-guided sampling during training and compare it to original random sampling (\cref{tab:sample_error}~IGR(curv) and IGR). Note that the point cloud inherits the curvature information from our initialized voxel grids since the points are extracted from the voxel grid. 
\par
\input{figures_tex/curvature_new.tex} 
\inparagraph{Off-Surface Point Sampling} Additionally, we show that our off-surface sampling method can be easily incorporated into previous works such as NeuralPull~\cite{ma2020neuralpull}. NeuralPull outperforms the other methods when normal information is not available. It uses Eq.~\cref{eq:neural_pull_loss} as the geometric loss to learn the pulling vector. However, it has to sample training points using nearest neighbor search and it fails easily when the points are too sparse or noisy. Our off-surface sampling strategy bypasses the nearest neighbor search and thus is more robust compared to the original NeuralPull. In~\cref{fg:neural_pull}, we show the result of our sampling method incorporated with NeuralPull loss compared to the original method together with the baseline method IGR. Our method and NeuralPull do not use normals in their loss terms, while IGR still uses point normals. The first two rows are synthetic datasets, and the last two rows are noisy real-world datasets \emph{vase} and \emph{sokrates}~\cite{zollhofer2015shading}. 
The noise originates from both the depth images and camera poses. IGR~\cite{gropp2020implicit} works well for synthetic datasets, whereas NeuralPull fails in sparse inputs. Our method compares favorably across all scenarios. Note that the key improvement is bypassing the nearest neighbor search step, thus for the works with similar step~\cite{ma2022reconstructing, li2022learning}, can easily employ our strategy. 

\subsection{Uncertainty Prediction}

\begin{figure}[t]
    \centering
    \begin{tabular}{m{0.2\linewidth}m{0.2\linewidth}m{0.2\linewidth}m{0.05\linewidth}m{0.2\linewidth}}
        \includegraphics*[width=0.9\linewidth, height=1.6cm]{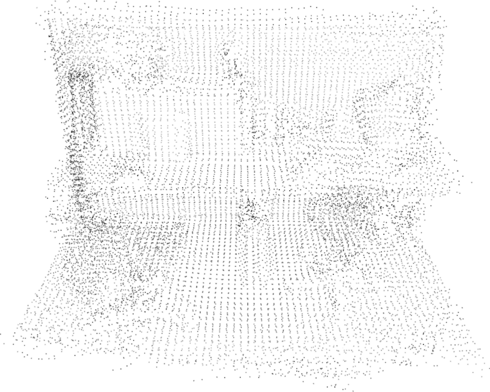} &
        \includegraphics*[width=0.9\linewidth, height=1.6cm]{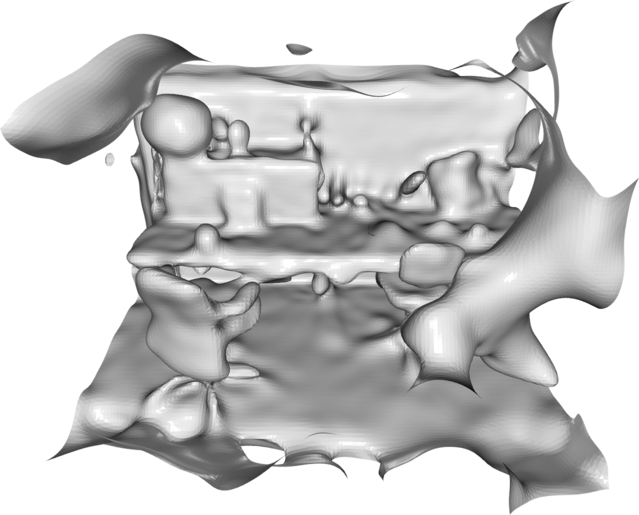} &
        \includegraphics*[width=0.9\linewidth, height=1.6cm]{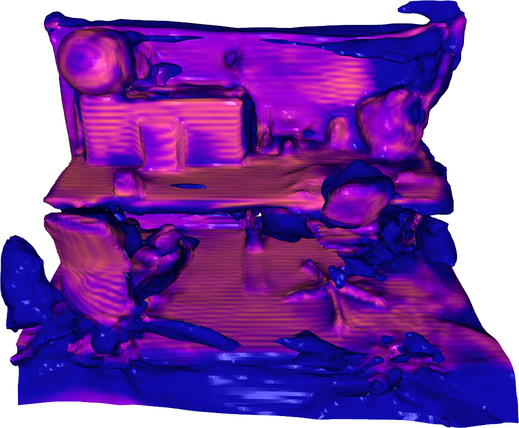} &
        \includegraphics*[height=1.6cm]{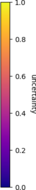}&
        \includegraphics*[width=0.9\linewidth, height=1.6cm]{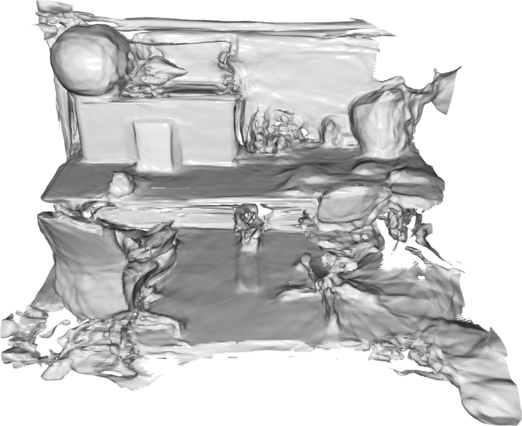}\\
        \centering Input & \centering SIREN & \centering Uncertainty & & {\centering Ours}\\
        \includegraphics*[width=\linewidth, angle=180, origin=c]{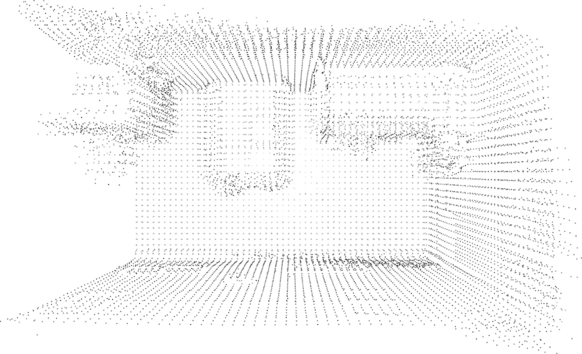} &
        \includegraphics*[width=\linewidth,angle=180, origin=c]{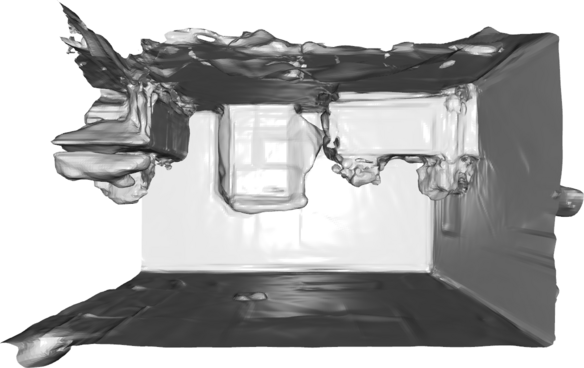} &
        \includegraphics*[width=\linewidth,angle=180, origin=c]{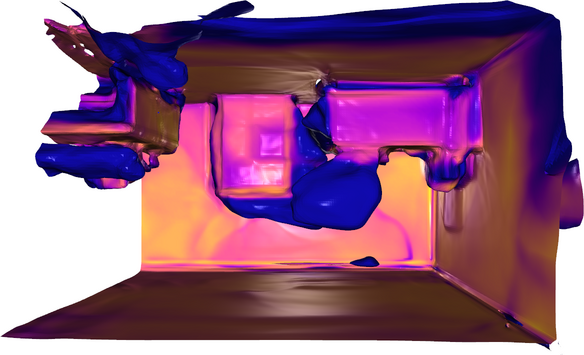} &
        \includegraphics*[height=1.5cm]{images/lr_kt0/64/error_bar_crop.png}&
        \includegraphics*[width=\linewidth, angle=180, origin=c]{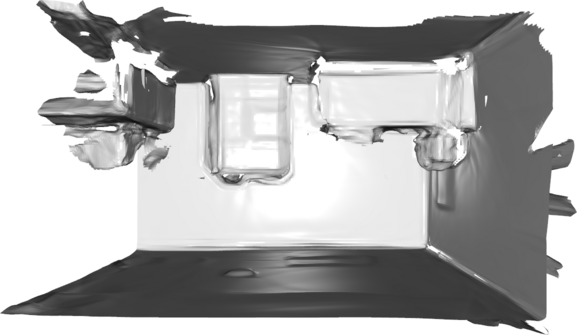}\\
        \centering Input & \centering IGR & \centering Uncertainty & & \centering Ours
    \end{tabular}
    \caption{Scene reconstruction results with sparse input on real-world dataset \emph{TUM\_rgbd} (first rows, sparse points $\sim 14k$), 
    with noisy camera poses. Synthetic dataset \emph{icl\_nium} (last rows, sparse points $\sim 14k$ )
    with ground truth camera poses. The complete comparison is included in supplementary material.}\label{fg:uncertainty}
 \vspace*{-0.6cm}
 \end{figure}

In this section, we show the uncertainty prediction result in~\cref{fg:uncertainty}, which illustrates that the uncertainty helps to eliminate redundant areas. We focus on showing the results on the scene dataset to show that with the help of uncertainty, we can also represent an open surface. The camera poses are estimated during the voxelization step. Many previous works~\cite{sitzmann2019siren,chibane2020ndf} have also trained neural networks to represent scene-level surfaces. However, a method such as~\cite{sitzmann2019siren} produces extra artifacts outside the surface. Although the authors propose one term in the loss function to penalize off-surface points for creating SDF values close to 0, it can not eliminate all artifacts, especially when the input is sparse and noisy. This problem can be solved by considering uncertainty during surface extraction as described in~\cref{subsec:our_method}. Due to space constraints, we show a subset of comparison results. For more results, please refer to the supplementary material.

\section{Conclusion}\label{sec:conclusion}

\inparagraph{Summary} In this work, we have presented novel curvature-guided sampling methods with uncertainty-augmented surface implicits representation.
Our method is easily transferred to existing methods that can efficiently deal with low-quality inputs. Our approach operates on depth images, which can be directly acquired from hardware. We propose a method that computes surface geometric properties: normals and curvatures on depth images instead of using ground truth mesh, which is more suitable for real-world applications.
The by-product uncertainty gives a reliability indication for the predicated signed distance value and can help with non-closed surface representations.

\inparagraph{Limitations and Future Work} Our approach does not specialize in surface completion. It will fail to recover any missing areas in depth. Next, since the presented method can be easily integrated with other neural reconstruction methods and shape completion techniques, we plan to incorporate shape completion and neural rendering to deal with missing areas.

%
%
\bibliographystyle{splncs04}
\bibliography{main}

\begin{thebibliography}{10}
\providecommand{\url}[1]{\texttt{#1}}
\providecommand{\urlprefix}{URL }
\providecommand{\doi}[1]{https://doi.org/#1}

\bibitem{atzmon2021augmenting}
Atzmon, M., Novotny, D., Vedaldi, A., Lipman, Y.: Augmenting implicit neural
  shape representations with explicit deformation fields. arXiv preprint
  arXiv:2108.08931  (2021)

\bibitem{de2000computational}
de~Berg, M., van Kreveld, M., Overmars, M., Schwarzkopf, O.: Computational
  Geometry: Algorithms and Applications. Springer-Verlag TELOS, Berlin,
  Heidelberg, 2 edn. (2000)

\bibitem{botsch2010polygon}
Botsch, M., Kobbelt, L., Pauly, M., Alliez, P., Levy, B.: Polygon Mesh
  Processing. CRC Press (2010)

\bibitem{Bylow2013Real}
Bylow, E., Sturm, J., Kerl, C., Kahl, F., Cremers, D.: Real-time camera
  tracking and 3d reconstruction using signed distance functions. In: Robotics:
  Science and Systems (2013)

\bibitem{Calakli2011}
Calakli, F., Taubin, G.: Ssd: Smooth signed distance surface reconstruction.
  Comput. Graph. Forum  \textbf{30}(7),  1993--2002 (2011)

\bibitem{carmo1976}
do~Carmo, M.P.: Differential geometry of curves and surfaces. Prentice Hall
  (1976)

\bibitem{chang2015shapenet}
Chang, A.X., Funkhouser, T., Guibas, L., Hanrahan, P., Huang, Q., Li, Z.,
  Savarese, S., Savva, M., Song, S., Su, H., Xiao, J., Yi, L., Yu, F.:
  Shapenet: An information-rich 3d model repository (2015)

\bibitem{chen2023local}
Chen, H., Zheng, C., Wampler, K.: Local deformation for interactive shape
  editing (2023)

\bibitem{chen2019learning}
Chen, Z., Zhang, H.: Learning implicit fields for generative shape modeling.
  In: 2019 IEEE/CVF Conference on Computer Vision and Pattern Recognition
  (CVPR). pp. 5932--5941 (2019)

\bibitem{chibane2020implicit}
Chibane, J., Alldieck, T., Pons{-}Moll, G.: Implicit functions in feature space
  for 3d shape reconstruction and completion. CoRR  \textbf{abs/2003.01456}
  (2020)

\bibitem{chibane2020ndf}
Chibane, J., Mir, A., Pons-Moll, G.: Neural unsigned distance fields for
  implicit function learning. In: Advances in Neural Information Processing
  Systems ({NeurIPS}) (December 2020)

\bibitem{redwood}
Choi, S., Zhou, Q.Y., Miller, S., Koltun, V.: A large dataset of object scans.
  arXiv:1602.02481  (2016)

\bibitem{bunny}
Curless, B., Levoy, M.: A volumetric method for building complex models from
  range images. In: Proceedings of the 23rd Annual Conference on Computer
  Graphics and Interactive Techniques. p. 303–312. SIGGRAPH '96, Association
  for Computing Machinery, New York, NY, USA (1996)

\bibitem{martino2014}
Di~Martino, J.M., Fernández, A., Ferrari, J.A.: 3d curvature analysis with a
  novel one-shot technique. In: 2014 IEEE International Conference on Image
  Processing (ICIP). pp. 3818--3822 (2014)

\bibitem{farin2002curves}
Farin, G.: Curves and Surfaces for CAGD: A Practical Guide. Computer graphics
  and geometric modeling, Elsevier Science (2002)

\bibitem{genova2019Deep}
Genova, K., Cole, F., Sud, A., Sarna, A., Funkhouser, T.A.: Deep structured
  implicit functions. ArXiv  \textbf{abs/1912.06126} (2019)

\bibitem{gropp2020implicit}
Gropp, A., Yariv, L., Haim, N., Atzmon, M., Lipman, Y.: Implicit geometric
  regularization for learning shapes. In: Proceedings of Machine Learning and
  Systems 2020, pp. 3569--3579 (2020)

\bibitem{iclnium}
Handa, A., Whelan, T., McDonald, J., Davison, A.: A benchmark for {RGB-D}
  visual odometry, {3D} reconstruction and {SLAM}. In: IEEE Intl. Conf. on
  Robotics and Automation, ICRA. Hong Kong, China (May 2014)

\bibitem{Hanocka_2020}
Hanocka, R., Metzer, G., Giryes, R., Cohen-Or, D.: Point2mesh: a self-prior for
  deformable meshes. ACM Transactions on Graphics  \textbf{39}(4) (Aug 2020)

\bibitem{hausdorff2005set}
Hausdorff, F.: Set Theory. American Mathematical Soc. (2005)

\bibitem{huang2023neural}
Huang, J., Gojcic, Z., Atzmon, M., Litany, O., Fidler, S., Williams, F.: Neural
  kernel surface reconstruction (2023)

\bibitem{Kazhdan2006poisson}
Kazhdan, M.M., Bolitho, M., Hoppe, H.: Poisson surface reconstruction. In:
  Sheffer, A., Polthier, K. (eds.) Proceedings of the Fourth Eurographics
  Symposium on Geometry Processing. SGP '06, vol.~256, pp. 61--70. Eurographics
  Association, Aire-la-Ville, Switzerland, Switzerland (2006)

\bibitem{kurita1999}
Kurita, T.: Computation of surface curvature from range images using
  geometrically intrinsic weights  (09 1999)

\bibitem{li2022learning}
Li, T., Wen, X., Liu, Y.S., Su, H., Han, Z.: Learning deep implicit functions
  for 3d shapes with dynamic code clouds (2022)

\bibitem{lorensen1987marching}
Lorensen, W.E., Cline, H.E.: Marching cubes: A high resolution 3d surface
  construction algorithm. ACM siggraph computer graphics  \textbf{21}(4),
  163--169 (1987)

\bibitem{ma2020neuralpull}
Ma, B., Han, Z., Liu, Y., Zwicker, M.: Neural-pull: Learning signed distance
  functions from point clouds by learning to pull space onto surfaces. CoRR
  \textbf{abs/2011.13495} (2020)

\bibitem{ma2022reconstructing}
Ma, B., Liu, Y.S., Han, Z.: Reconstructing surfaces for sparse point clouds
  with on-surface priors (2022)

\bibitem{lars2018occupancy}
Mescheder, L.M., Oechsle, M., Niemeyer, M., Nowozin, S., Geiger, A.: Occupancy
  networks: Learning 3d reconstruction in function space. CoRR
  \textbf{abs/1812.03828} (2018)

\bibitem{Michalkiewicz2019deep}
Michalkiewicz, M., Pontes, J.K., Jack, D., Baktashmotlagh, M., Eriksson, A.P.:
  Deep level sets: Implicit surface representations for 3d shape inference.
  CoRR  \textbf{abs/1901.06802} (2019)

\bibitem{newcombe2011}
Newcombe, R.A., Izadi, S., Hilliges, O., Molyneaux, D., Kim, D., Davison, A.J.,
  Kohi, P., Shotton, J., Hodges, S., Fitzgibbon, A.: Kinectfusion: Real-time
  dense surface mapping and tracking. In: 2011 10th IEEE international
  symposium on mixed and augmented reality. pp. 127--136. Ieee (2011)

\bibitem{novello2022exploring}
Novello, T., Schardong, G., Schirmer, L., da~Silva, V., Lopes, H., Velho, L.:
  Exploring differential geometry in neural implicits (2022)

\bibitem{park2019deepsdf}
Park, J.J., Florence, P., Straub, J., Newcombe, R.A., Lovegrove, S.: Deepsdf:
  Learning continuous signed distance functions for shape representation. CoRR
  \textbf{abs/1901.05103} (2019)

\bibitem{peng2021shape}
Peng, S., Jiang, C.M., Liao, Y., Niemeyer, M., Pollefeys, M., Geiger, A.: Shape
  as points: A differentiable poisson solver (2021)

\bibitem{richa2022unsigned}
Richa, J.P., Deschaud, J.E., Goulette, F., Dalmasso, N.: Unsigned distance
  field as an accurate 3d scene representation for neural scene completion
  (2022)

\bibitem{sitzmann2019siren}
Sitzmann, V., Martel, J.N., Bergman, A.W., Lindell, D.B., Wetzstein, G.:
  Implicit neural representations with periodic activation functions. In: Proc.
  NeurIPS (2020)

\bibitem{sommer2022}
Sommer, C., Sang, L., Schubert, D., Cremers, D.: Gradient-{SDF}: {A}
  semi-implicit surface representation for 3d reconstruction. In: IEEE
  Conference on Computer Vision and Pattern Recognition (CVPR) (2022)

\bibitem{spivak1999comprehensive}
Spivak, M.: A Comprehensive Introduction to Differential Geometry. No. Bd. 1 in
  A Comprehensive Introduction to Differential Geometry, Publish or Perish,
  Incorporated (1999)

\bibitem{sturm12iros}
Sturm, J., Engelhard, N., Endres, F., Burgard, W., Cremers, D.: A benchmark for
  the evaluation of rgb-d slam systems. In: Proc. of the International
  Conference on Intelligent Robot Systems (IROS) (Oct 2012)

\bibitem{weder2020routfusion}
Weder, S., Sch{\"{o}}nberger, J.L., Pollefeys, M., Oswald, M.R.: Routedfusion:
  Learning real-time depth map fusion. CoRR  \textbf{abs/2001.04388} (2020)

\bibitem{yang2021geometry}
Yang, G., Belongie, S., Hariharan, B., Koltun, V.: Geometry processing with
  neural fields. In: Thirty-Fifth Conference on Neural Information Processing
  Systems (2021)

\bibitem{zollhofer2015shading}
Zollh{\"o}fer, M., Dai, A., Innmann, M., Wu, C., Stamminger, M., Theobalt, C.,
  Nie{\ss}ner, M.: Shading-based refinement on volumetric signed distance
  functions. ACM Transactions on Graphics (TOG)  \textbf{34}(4),  1--14 (2015)

\end{thebibliography}

\newpage
\appendix

\section{Appendix}\label{sec:appendix}

\renewcommand{\thefigure}{A.\arabic{figure}}
\renewcommand{\thetable}{A.\arabic{table}}
\renewcommand{\thesection}{A\arabic{section}}

\section{Code, Datasets and baseline methods}

Our code and evaluation scripts will be publicly available upon acceptance. We will also provide the detailed information about the code of baseline methods.

\begin{table}[h]
    \centering
    \begin{tabular}{l p{2cm} l l p{5cm} p{3cm}}
    \toprule
         & name & type & year & link & license \\
    \midrule
        \cite{redwood} & Redwood & dataset & 2016 & {\small\url{http://www.redwood-data.org/3dscan/}} & Public Domain \\
        \cite{bunny} & The Stanford 3D & dataset & 1994 & {\small\url{http://graphics.stanford.edu/data/3Dscanrep/}} & Public Domain \\
        \cite{zollhofer2015shading} & multi-view dataset & dataset & 2015 & {\small\url{http://graphics.stanford.edu/projects/vsfs/}} & CC BY-NC-SA 4.0 \\
        \cite{iclnium} & ICL-NUIM & dataset & 2014 & {\small\url{https://www.doc.ic.ac.uk/~ahanda/VaFRIC/iclnuim.html}} & CC BY 3.0 \\
        \cite{sturm12iros} & TUM-rgbd & dataset & 2012 & {\small\url{https://cvg.cit.tum.de/data/datasets/rgbd-dataset}} & CC BY 4.0 \\
        \cite{sommer2022} & gradient-SDF & code & 2022 & {\small\url{https://github.com/c-sommer/gradient-sdf}} & BSD-3 \\
        \cite{gropp2020implicit} & IGR & code & 2020 & {\small\url{https://github.com/amosgropp/IGR}} & - \\
        \cite{sitzmann2019siren} & SIREN & code & 2019 & {\small\url{https://github.com/vsitzmann/siren}} & MIT license \\
        \cite{chibane2020implicit} &IF-NET & code & 2020 & {\small\url{https://virtualhumans.mpi-inf.mpg.de/ifnets/}} & - \\
        \cite{ma2020neuralpull} & Neural-Pull & code & 2021 & {\small\url{https://github.com/bearprin/neuralpull-pytorch}} & - \\
        \cite{chibane2020ndf} & NDF & code & 2020 & {\small\url{https://virtualhumans.mpi-inf.mpg.de/ndf/}} & -\\
        \cite{Kazhdan2006poisson} & Poisson & code & 2006 & {\small\url{http://www.open3d.org/}} & - \\
        \cite{Calakli2011} & SSD & code & 2011 & {\small\url{http://mesh.brown.edu/ssd/software.html}} & - \\
    \bottomrule
    \label{tab:code_data}
    \end{tabular}
      \caption{Used datasets and code in our submission, together with reference, link, and license. We did our real-world experiments on two datasets, multi-view dataset~\cite{zollhofer2015shading} (for which ground truth poses exist), and Redwood~\cite{redwood} (without ground truth poses). Two synthetic dataset, the Stanford 3D~\cite{bunny}, which is an object dataset, and ICL-NUIM dataset~\cite{iclnium}, which is a scene dataset. For the comparison methods, we use the code listed in the table.}
    \vspace{0.25cm}
\end{table}

This Supplement contains information on code, datasets, and more comparison results. The citation numbers are the same as in the main paper. \textbf{Our code and evaluation scripts will be publicly available upon acceptance.}

\section{Mathematical detail}
\subsection{Math Notations}
We summarize important math notation we used in the paper and appendix in Table~\cref{table:notation}.

\begin{longtable}{lp{4.2cm}|lp{4.2cm}}
 \toprule
        \textbf{Symbol} & \textbf{Description} & \textbf{Symbol} & \textbf{Description} \\
		\toprule
		  $\vt{x}\in\mathbb{R}^3$ & 3D points & $\mathcal{P}\subset\mathbb{R}^3$ & point cloud set\\
		  $\mathcal{V}\subset\mathbb{N}^+$ & points index set & $\surf{S}\subset\mathbb{R}^3$ &continuous surface\\
		  $f(\vt{x}, \theta)$ & neural implicit function &  $\theta\in\mathbb{R}^{n\times m}$ &learnable parameter\\
        $w^p \in [0,1]$ & points uncertainty & $w^v \in [0,1]$ & voxel uncertainty \\
		  $\psi^v\in\mathbb{R}$ & voxel SDF value & $\psi^p\in\mathbb{R}$ & point SDF value \\
        $\hat{\vt{g}}^v\in\mathbb{R}^3$ &  normalized distance gradient & $\vt{g}^v\in\mathbb{R}^3$ & voxel distance gradient\\
        $\nabla$ & differential operator & $H^p\in\mathbb{R}$ & point mean curvature \\
         $\Gamma\subset\mathbb{R}^3$ & sample domain & $\gamma\in\mathbb{R}$ & SDF threshold \\
         $K\in\mathbb{R}$ & Gaussian curvature & $H\in\mathbb{R}$ & mean curvature \\
         $k_1$, $k_2$ & principal curvature & $D\subset\mathbb{R}^2$ & depth image\\
         $\Omega\subset\mathbb{R}^2$ & image domain & $\mathcal{M}\subset\mathbb{R}^3$ & Monge path \\
         $d_\surf{S}(\cdot)$ & signed distance to surface $\surf{S}$ & $\Gamma^{+}\subset\mathbb{R}^3$ & sample domain with positive uncertainty \\
         $\bar{H}\in\mathbb{R}$ & higher threshold & $\underbar{H}\in\mathbb{R}$ & lower threshold of curvature \\
         $\vt{n}_i\in\mathbb{R}^3$ & known points normal & $\mat{Q}\in\mathbb{R}^{3\times 3}$ & camera intrinsic matrix \\
         $\mat{R}\in SO(3)$ & camera rotation matrix & $\vt{t}\in\mathbb{R}^3$ & camera translation vector \\ 
		\bottomrule\\
	\caption{Summary of our notation in the main paper and the supplementary material.}
	\label{table:notation}
\end{longtable}

\subsection{Voxelization Details}\label{subsec:s_voxelization}
Given an incoming depth $D(m,n), (m,n)\in\Omega$ with $z = D(m,n)\in\mathbb{R}$ and the estimated pose $\mat{R}, \vt{t}$, the 3D points in world coordinates are 
\begin{align}
    \vt{x} &= \mat{R} \mat{Q}^{-1} \begin{bmatrix}m \\ n \\ 1 \end{bmatrix} z \,, \label{eq:re-project} \\ 
    \mat{Q} &= \begin{bmatrix} f_x & 0 & c_x \\ 0 & f_y & c_y \\ 0& 0& 1\end{bmatrix} \label{eq:intrinsic} \,,
\end{align}
where $Q$ is the camera intrinsic matrix. Here we describe the standard voxel integration procedure. For each voxel $\vt{v}_i$, we project voxel center to the current depth image using $\vt{p} = (p_x, p_y, p_z) = \mat{R}^\top(\mat{Q}\vt{v}_i - \vt{t})$ and $(m, n, 1) = \vt{p}/p_z$ to find the corresponding pixel coordinates $(m,n)$ on depth, then compare the $z-axis$ value, which stands for the distance of voxel $\vt{v}_i$ to the camera center. Then we compute the difference of $q_z$ with depth value on the corresponding pixel, then the project is to the normal direction to compute the point-to-plan distance (see~\cref{eq:distance}). We use the convention that inside the surface is positive and the outside surface is negative, such that the gradient of SDF has the same sign of surface normal. Then, the uncertainty of each voxel is computed using the distance $d_{\mathcal{S}}(\vt{v}_i)$. For visible voxel from current depth, which means $d_{\mathcal{S}}(\vt{v}_i)$ is negative, the uncertainty is set to $1$, which means this update of distance is valid. To allow some noise, we set the depth integration threshold $T$ (we use $T=5$) as a truncate threshold. If $d_{\mathcal{S}}(\vt{v}_i)$ is positive and smaller $Tv_s$, then the uncertainty drops linearly to $0$ (see~\cref{eq:uncertainty}). Written in more simplified mathematical expression is, the SDF value of points $\vt{x}_j$ and its normal $\vt{n}_j$ and curvature $H_j$ computed from the depth image $D$, then the voxel grid $\{\vt{v}_i\}$ SDF and uncertainty is computed by
\begin{align}
    & d_{\surf{S}}(\vt{v}_i) = (\vt{x}_{j^*} - \vt{v}_i)^\top \hat{\vt{g}}_i \label{eq:distance} \\
    & \nabla d_{\surf{S}}(\vt{v}_i) = \vt{g}_i = \mat{R} \vt{n}_{j^*} \\
    & w^v(\vt{v}_i) = \begin{cases}
        1, ~~ d_{\surf{S}}(\vt{v}_i) \leq 0 \\
        1 - \frac{d_{\surf{S}}(\vt{v}_i)}{v_s T}, ~~ 0 < d_{\surf{S}}(\vt{v}_i) \leq  v_sT\\
        0, ~~ \text{else}
    \end{cases} \label{eq:uncertainty} \\
    & j^* = \argmin_j \norm{\vt{x}_j - \vt{v}_i}
    \end{align}
where $v_s$ is the voxel size, $T\in\mathbb{N}^+$ is the truncate voxel number, in this paper, voxel size is set to $0.8cm$ for $64^3$ grid and $0.2cm$ for $256^3$ grid with $T=5$. Iterating over all depth image, the SDF $\psi^v_i$, uncertainty $w^v_i$, gradient $\vt{g}_i$ and curvature $H^v_i$ is updated by
\begin{align}
\psi^v_i & \longleftarrow \frac{w^v_i \psi^v_i + w^v(\vt{v}_i)d_{\surf{S}}(\vt{v}_i) }{w^v_i + w^v(\vt{v}_i)} \\
\vt{g}_i &\longleftarrow \frac{w^v_i\vt{g}_i^v + w^v(\vt{v}_i)\mat{R}\vt{g}^v_i}{w^v_i + w^v(\vt{v}_i)}\\
H^v_i &\longleftarrow \frac{w^v_i K^v_i + w^v(\vt{v}_i)K_{j^*}}{w^v_i + w^v(\vt{v}_i)}\\
w^v_i & \longleftarrow w^v_i + w^v(\vt{v}_i)
\end{align}
The attributes in voxels are integrated using weighted averages to be more robust to the noise, especially considering that the normal and curvature are computed using neighborhood information of one pixel on the depth map.

\subsection{Camera pose estimation}
We can estimate the camera pose when we initialize the voxel grid using the same method as Gradient-SDF~\cite{sommer2022}. We use the first depth map with $\mat{R}_0 = \mat{I}$ and $\vt{t}_0 = \bm{0}$ to initial voxel grid as described before in~\cref{subsec:s_voxelization}, then the initial surface $\mathcal{S}$ in contained in voxels. Then from the second depth, each depth, \eg depth $k$ is an incoming point cloud $\mathcal{P}^k = \{\vt{p}^k_j\}_j$ (outside index means iterate over $j$) after we project them using~\cref{eq:re-project}. The problem then is convert to find $\mat{R}_k$, $\vt{t}_k$, such that 
\begin{equation}
    (\mat{R}_k, \vt{t}_k) = \argmin_{\substack{\mat{R}\in SO(3), \\ \vt{t} \in \mathbb{R}^3}} \sum_j w_j d_{\mathcal{S}}(\mat{R}_k\vt{p}^k_j+\vt{t}_k) \,,\label{eq:camera_loss}
\end{equation}
where $SO(3)$ is 3-dimensional Lie group and $d_{\mathcal{S}}(\vt{p})$ denotes the signed distance from the point $\vt{p}$ to surface $\mathcal{S}$
\begin{equation}
    |d_{\mathcal{S}}(\vt{p})| = \min_{\vt{p_s}\in\mathcal{S}} \norm{\vt{p} -\vt{p}_s}\,.
\end{equation}
~\cref{eq:camera_loss} is solved by Gaussian-Newton and since the gradient $\nabla d_{\mathcal{S}}(\vt{p}) = \hat{\vt{g}}^v$ is pre-computed and stored, it also accelerates the optimization steps, as described in~\cite{sommer2022}. The camera pose is not given for Redwood~\cite{redwood} (sofa, kiosk, washmachine sequences) and TUM\_RGBD~\cite{sturm12iros} (household) datasets, the poses are estimated as described in this section.

\subsection{Proof of the Curvature Integration}
In the paper section~\cref{subsec:voxelization}, we mention that the transformation between two depth coordinates has non-zero Jacobian (non-zero determinant); hence, integrating curvatures from depth images makes sense. Here is the formulation and proof.

\textit{The determinant of the Jacobian of the parameter transformation of the parameterization in the two depth images is non-zero; the mean curvature $H(x,y)$ and Gaussian curvature $K(x,y)$ are invariant.}

\begin{proof}
    Given two depth images $D_1$ and $D_2$ taken at two different positions. Suppose the transformation from position $1$ to position $2$ is a rigid body motion $\mat{T} = [\mat{R}, \vt{t}]$, where $\mat{R}\in SO(3)$ is a rotation matrix and $\vt{t} \in \mathbb{R}^3$ is a translation vector. Let pixel $p=(m,n)$ in $D_1$, $0\neq z = D_1(m,n)$, and $\mat{Q}\in\mathbb{R}^{3\times 3}$ be the  camera intrinsic matrix, then the transformation of pixel $p$ to $\bar{p} = (\bar{x}, \bar{y})$ in $D_2$ is
    \begin{align}
        \bar{z} \begin{bmatrix}\bar{x} \\ \bar{y} \\ 1 \end{bmatrix} = \mat{Q} \vt{x} \,,
    \end{align}
    where $\vt{x}$ is computed using~\cref{eq:re-project} and $\mat{Q}$ is same as~\cref{eq:intrinsic}.
    is invertible, $f_x$, $f_y$ is the camera focal length and $c_x$, $c_y$ is the principal points. The pixel $3D$ coordinates in under two camera view is $\mat{Q}^{-1}(x,y,1)^\top z$ and $\mat{Q}^{-1}(\bar{x}, \bar{y}, 1)^\top \bar{z}$.
    Let $\{r_{ij} \}_{ij}, i,j \in \{1,2,3\}$ be the element in $\mat{R}$ and $\vt{t} = (t_1, t_2, t_3)^\top$, compute the right side, we have $\bar{z} = r_{31}x + r_{32}y + r_{33}z + t_3$ is the depth value after a rigid body motion and  $\bar{z}\neq 0$ since it does not fall to image plane of $D_2$ as we assume the point is visible in both camera position. 
    \begin{equation}
        \det (\mat{Q}\mat{R}\mat{Q}^{-1}) = \det(\mat{Q})\det(\mat{R})\det(\mat{Q})^{-1} = 1\,,
    \end{equation}
    
    it is because $\det(\mat{Q}) \neq 0$,  $\det (\mat{R}) = 1$ and $\det(\mat{Q}^{-1}) = \det(\mat{Q})^{-1}$. Thus, the transformation Jacobian of the 3D points is non-zero. For the Jacobian of the transformation from $(x,y)$ to $(\bar{x}, \bar{y})$, we only need to consider the upper-left $2\times 2$ submatrix of $\mat{Q}\mat{R}\mat{Q}^{-1}$. Since the upper-left $2\times 2$ matrix of $\mat{R}$ represents the rotation matrix in the $xy$ plane, thus it is non-zero. The upper-left $2\times 2$ matrix of $\mat{Q}$ is diagonal also non-zero.  Hence, we get the determinant of Jacobian from $(x,y)$ to $(\bar{x}, \bar{y})$ is also non-zero.
\end{proof}

\subsection{Main Curvature Vs. Gaussian Curvature}
In this paper, we focus only on the main curvature and have not provided an ablation study to compare the influence of two curvatures. The reasons are the following. First, from the curvature visualization (\cref{fg::curvature} and~\cref{fg::sample_points}), the distribution of low, median, and high main curvature and Gaussian curvature are more or less similar. Since we only need the curvature as a guide information to sample points, by controlling the low, median, and high curvature threshold and proportion of each category, using main curvature and Gaussian curvature should not make much difference. Second, from a mathematical perspective, main curvature is the addition of two principal curvatures, \ie $H = \frac{k_1 + k_2}{2}$ and Gaussian curvature is $K = k_1k_2$. We can have the following bound 
$K \leq H^2$.
Thus, the difference between using the main curvature and the Gaussian curvature is negligible. 

\section{Depth-based Method Comparison}
\begin{figure}[t]
    \centering
       \begin{tabular}{cccccccc}
        \includegraphics[width=0.12\linewidth]{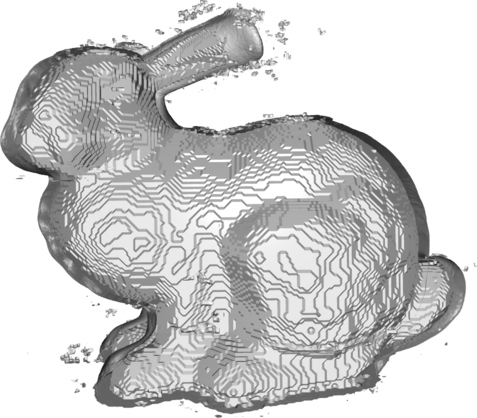} &
        \includegraphics[width=0.12\linewidth]{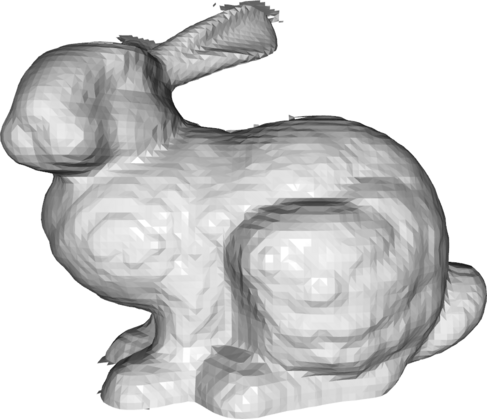} &
        \includegraphics[width=0.12\linewidth]{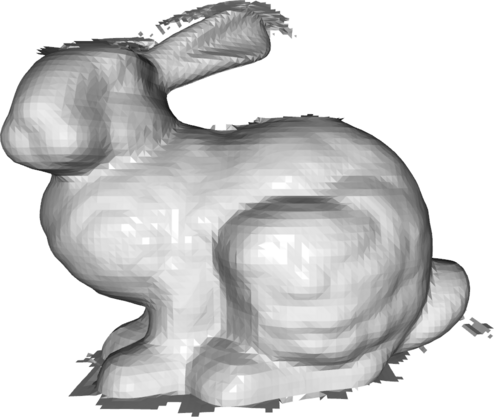} &
        \includegraphics[width=0.12\linewidth]{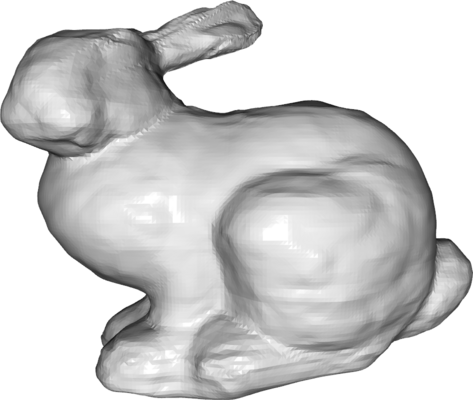} &
        \includegraphics[width=0.12\linewidth]{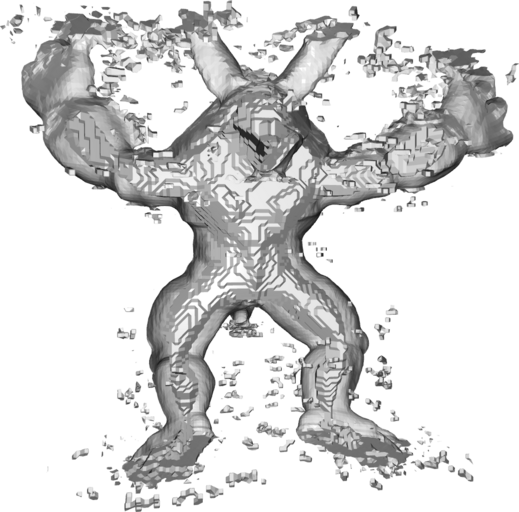} &
        \includegraphics[width=0.12\linewidth]{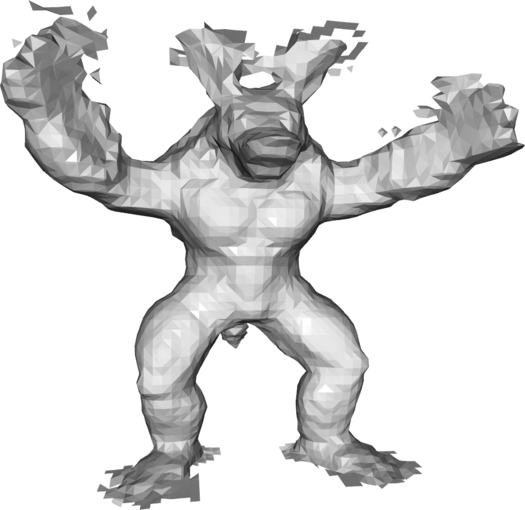} &
        \includegraphics[width=0.12\linewidth]{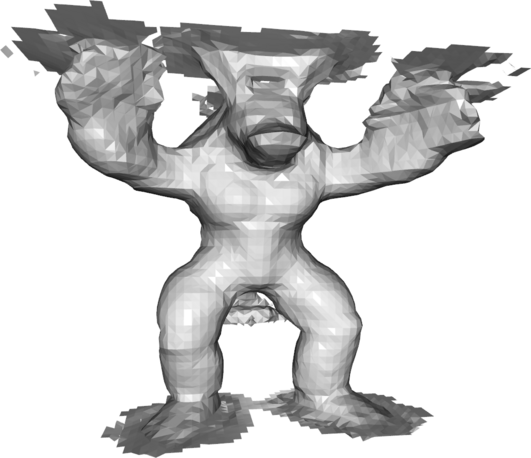} &
        \includegraphics[width=0.12\linewidth]{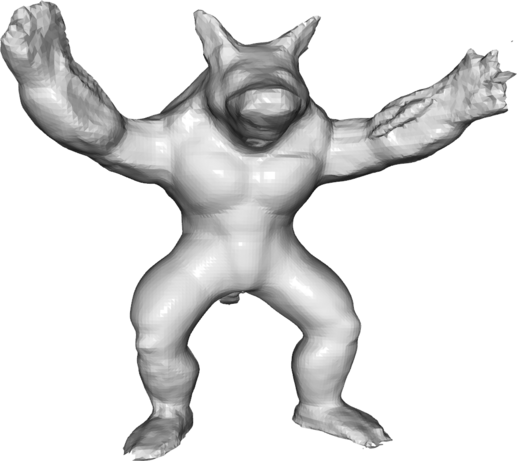} \\
        \includegraphics[height=0.12\linewidth, angle=90, origin=c]{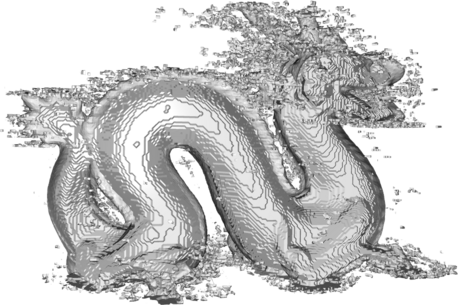} &
        \includegraphics[height=0.12\linewidth, angle=90, origin=c]{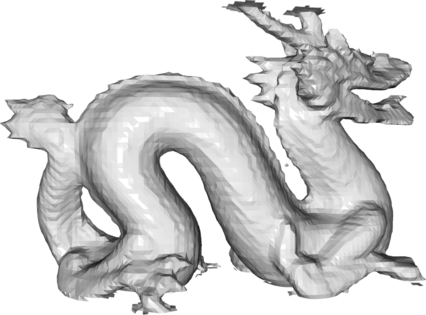} &
        \includegraphics[height=0.12\linewidth, angle=90, origin=c]{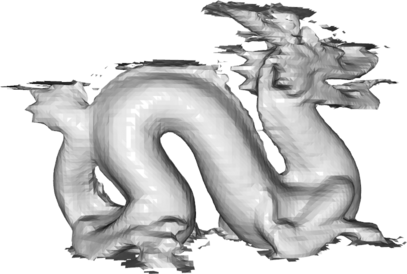} &
        \includegraphics[height=0.12\linewidth, angle=90, origin=c]{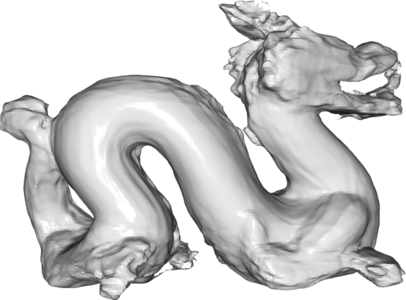} &
        \includegraphics[width=0.08\linewidth]{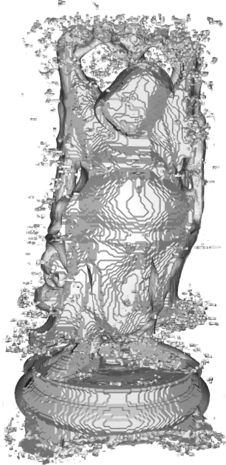} &
        \includegraphics[width=0.08\linewidth]{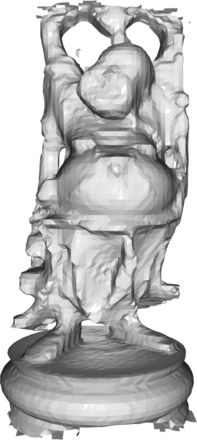} &
        \includegraphics[width=0.08\linewidth]{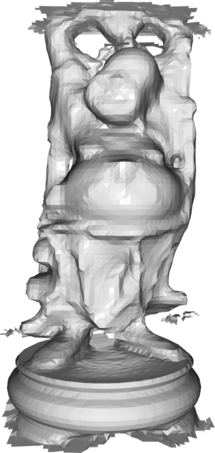} &
        \includegraphics[width=0.08\linewidth]{images/happy_budda/64/ours_crop.png} \\
        RoutedFusion & TSDF & Grad-SDF & Ours & RoutedFusion & TSDF & Grad-SDF & Ours
       \end{tabular}
       \caption{The visualization of the reconstructed meshes on synthetic datasets. Compared with depth based methods: RoutedFusion~\cite{weder2020routfusion}, TSDF~\cite{Bylow2013Real}, Gradient-SDF (Grad-SDF)~\cite{sommer2022} and our method. TSDF uses $128^3$ resolution voxel grid with $v_s=0.02$ (m) voxel size, Grad-SDF uses $v_s=0.02$ (m), ours is built on coarse voxel grid with $64^3$ resolution voxel with $v_s=0.04$ (m). }\label{fg::depth_compare_synthetic}
       \vspace{-0.5cm}  
       \end{figure}

\begin{figure}[t]
        \centering
           \begin{tabular}{cccc}
            \includegraphics[width=0.22\linewidth]{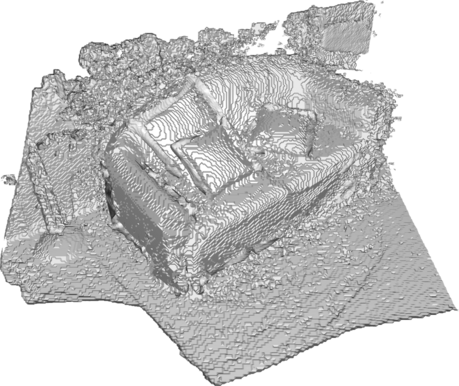} &
            \includegraphics[width=0.22\linewidth]{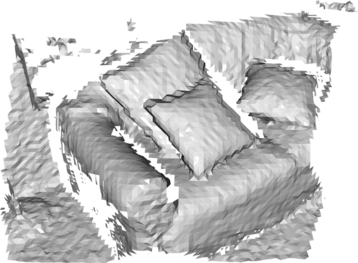} &
            \includegraphics[width=0.22\linewidth]{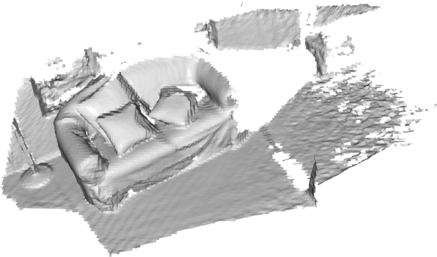} &
            \includegraphics[width=0.22\linewidth]{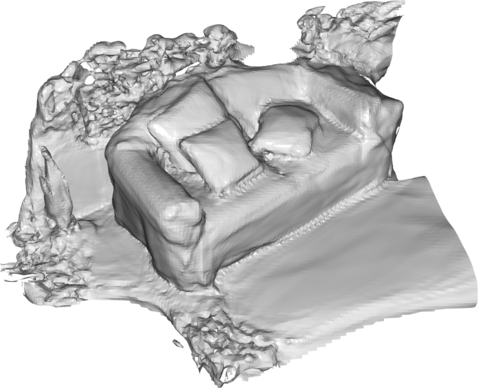} \\
            \includegraphics[width=0.22\linewidth]{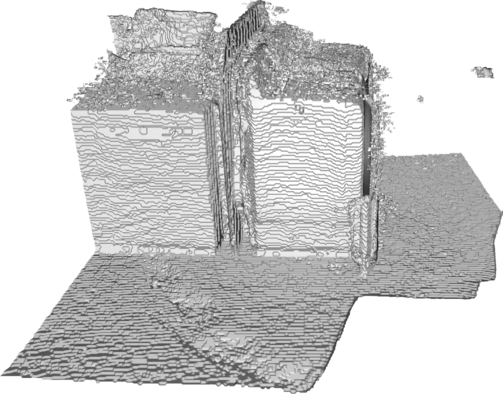} &
            \includegraphics[width=0.22\linewidth]{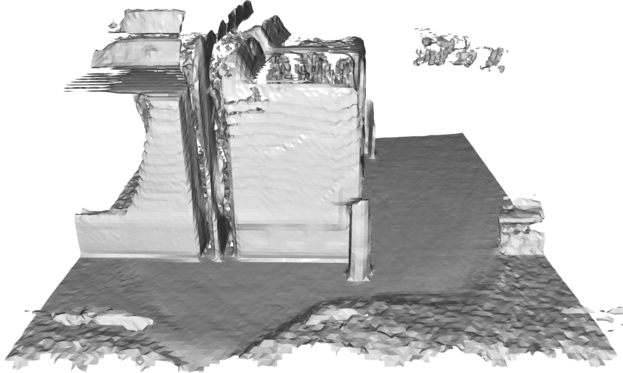} &
            \includegraphics[width=0.22\linewidth]{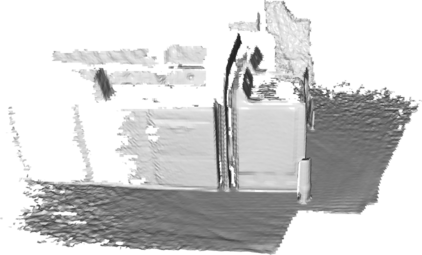} &
            \includegraphics[width=0.22\linewidth]{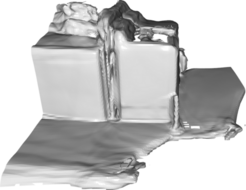} \\
            RoutedFusion & TSDF & Grad-SDF & Ours
           \end{tabular}
\caption{The Visualization of reconstructed meshes on real world datasets. For Gradient-SDF (Grad-SDF)~\cite{sommer2022}, TSDF~\cite{Bylow2013Real} and our voxelization, the camera poses are estimated during depth-fusion, RoutedFusion~\cite{weder2020routfusion} takes estimated camera poses from our methods. The voxel setting is same as synthetic datasets.}\label{fg::depth_compare_real}
        \end{figure}

We show more comparison results with the depth-based methods. All time evaluations are done on a computer with Intel Xeon(R) CPU @ 3.60GHz and 12 GeForce GTX TITAN X GPU. \cref{fg::depth_compare_synthetic} shows the visualization results of synthetic datasets and~\cref{fg::depth_compare_real} shows the results on real-world datasets. \par
\textbf{RoutedFusion}~\cite{weder2020routfusion} provides networks trained for depth fusion across datasets. Ground truth or pre-generated mesh and camera poses are needed during training. For synthetic datasets, ground truth meshes and camera poses are given. For real-world datasets (\eg~\cite{redwood}), pre-estimated camera poses using our pipeline and pre-generated mesh using TSDF~\cite{newcombe2011} are given. The training time for two networks: Rout and Fusion needs around 70 hours for 360 ($640\times480$) images from 4 synthetic datasets for each network. We tested on their pre-trained model and our trained model, pre-trained model offers better (less noisy) results. Thus we show the fusion results from their pre-trained model. During inference, ground truth poses and meshes are still needed for a guided fusion. The inference time for fusion is around 1.5 seconds per frame with CUDA. As shown in~\cref{fg::depth_compare_synthetic} and~\cref{fg::depth_compare_real}, RoutedFusion~\cite{weder2020routfusion} tends to still perseve contour of voxels the same as IF-NET~\cite{chibane2020implicit} (see~\cref{fg:sampling_visual} in main paper and~\cref{fg:appendix_visual} in supplementary). Plus it creates noise around the meshes. \par
\textbf{TSDF}~\cite{newcombe2011} we re-implemented TSDF tracker according to~\cite{Bylow2013Real}. The method builds a dense voxel grid for tracking and depth fusion. The code is implemented purely on CPU, and each voxel store a signed distance and a weight. We use a $128^3$ resolution grid with $2cm$ voxel size, which need $16$ Mb memory. For each depth, all voxels are project to image plane to compare with the depth value. Thus, the running time dependents on voxel resolution and depth image size. Fusion time is around $23ms$ per frame ($640\times480$). TSDF has a limited reconstruction range, which equals resolution times the voxel size. Because of the truncation and missing data in depth images, it might create holes on the surface.\par
\textbf{Gradient-SDF}~\cite{sommer2022} uses a hash-map to store sparse voxels.  For each voxel it stores a signed distance, a weight and a gradient of the signed distance. Different from traditional volumetric grid methods, instead of updating every pre-defined voxel, it updates according to each pixel in the incoming depth image, \ie, for each incoming depth image, points are unprojected to 3D coordinates, it then computes which voxels need to be updated according to these points. Thus, the running time per frame depends on depth image size solely. However, it also means that for each depth, the number of updated voxels is at most equal to the pixel number of the depth image. The memory consumption of Gradient-SDF is dynamical, depends on each datasets and how many voxels are created during the fusion, (\eg, for bunny the memory consumption is $4.4$ Mb and for Armadillo is around $1.5$ Mb.)  The fusion time is around $30ms$ per frame ($640\times480$). One possible \par
\textbf{Our voxelization} builds a dense voxel grid with resolution $64^4$, stores a signed distance, a gradient of signed distance, an uncertainty, and curvature of each voxel. The memory consumption is $6$ Mb. The fusion time is around $31ms$ per frame.

\setlength{\tabcolsep}{0pt}

\section{Loss Ablation Study}
\inparagraph{Training details} We use an 8-layer multi-layer perceptron (MLP) with ReLU activations. Each layer has $256$ nodes, and the last layer has $2$ output nodes for the SDF and uncertainty. We set the learning rate to $10^{-4}$ with decay. The batch size is $10k$, and we train for $10k$ epochs for each dataset. Our PyTorch implementation takes approximately 15 minutes to train on a GeForce GTX TITAN X GPU with CUDA for each dataset. The setting is the same for all the experiments for the proposed method.\par
Our loss function is composed of four distinct terms, and in this section, we detail how each contributes to the overall results. Table~\cref{tab:ablation_error} presents a quantitative assessment of the reconstructed meshes, indicating the impact of each loss term. The Eikonal term~\cref{eq:eikonal}, ensures that our neural network functions as a signed distance field. The geometric loss term~\cref{eq:geo_loss} is key in clarifying the orientation of surfaces. The uncertainty term~\cref{eq:weight} helps eliminate undesired areas. Furthermore, we investigate the effects of off-surface sampling in scenarios lacking normal information. Although our process incorporates both normal and curvature information—deriving the normal from depth—we still explore the role of normal information in processing sparse and dense inputs. This is to demonstrate that point cloud density and normal information jointly contribute to resolving surface orientation challenges, as shown in~\cref{tab:ablation_igr}, where we compare the error metrics of our method against IGR~\cite{gropp2020implicit} with and without the geometric loss term. Interestingly, for certain datasets, omitting the geometric loss~\cref{eq:geo_loss} or Eikonal loss~\cref{eq:eikonal} leads to reduced error rates. This suggests that at certain points, the model struggles to simultaneously satisfy both constraints, according to our analysis.

\begin{table}[t]
    \centering
    \scriptsize
    \begin{tabular}{|p{1cm}|p{1.3cm}|ccccc|ccccc|}
        \toprule
        \multirow{2}{*}{\textbf{Metric}} & \multirow{2}{*}{\textbf{Dataset}} & \multicolumn{5}{c|}{\textbf{Method} (Sparse)} & \multicolumn{5}{c|}{\textbf{Method} (Dense)} \\
        \cline{3-12}
         & & w/o $\mathit{l}_{\mathcal{E}}$ & w/o $\mathit{l}_{\mathcal{W}}$ & w/o $\mathit{l}_{\mathcal{N}}$ & w/o $\mathit{l}_{\mathcal{N}, \mathcal{W}}$ & Ours & w/o $\mathit{l}_{\mathcal{E}}$ & w/o $\mathit{l}_{\mathcal{W}}$ & w/o $\mathit{l}_{\mathcal{N}}$ & w/o $\mathit{l}_{\mathcal{N}, \mathcal{W}}$ & Ours\\
        \midrule
        \multirow{4}{*}{\begin{minipage}{1cm}\textbf{CD}\\($\times 10^2$)\end{minipage}} & Bunny & 0.152 & 0.176 & 0.171 & 0.265 & \textbf{0.067} & 0.153 & 0.139 & 0.138 & 0.395 & \textbf{0.068} \\
         & Armadillo & 0.089 & 0.052 & \textbf{0.037} & 1.180 & \textbf{0.037} & 0.104 & 0.041 & 0.052 & 1.987 & \textbf{0.030} \\
         & Dragon & 0.387 & 0.182 & 0.226 & 0.251 & \textbf{0.157} & 0.138 & 0.158 & 0.184 & 0.196 & \textbf{0.126} \\
         & Buddha & 0.378 & 0.289 & 0.467 & 0.184 & \textbf{0.125} & 0.248 & 0.301 & 0.253 & 0.295 & 0.267\\  
        \midrule
        \multirow{4}{*}{\textbf{HD}} & Bunny & 0.058 & 0.044 & 0.039 & 0.074 & \textbf{0.016} & \textbf{0.010} & 0.139 & 0.012 & 0.039 & 0.043 \\
        & Armadillo & 0.009 & 0.016 & \textbf{0.006} & 0.104 & \textbf{0.006} & 0.019 & 0.013 & 0.028 & 0.251 & \textbf{0.007} \\
        & Dragon & 0.044 & 0.038 & 0.048 & 0.047 & \textbf{0.030} & 0.024 & 0.045 & 0.065 & 0.041 & \textbf{0.025} \\
        & Buddha & 0.118 & 0.057 & 0.146 & 0.045 & \textbf{0.017} & 0.052 & 0.078 & 0.053 & 0.142 & \textbf{0.044} \\
        \bottomrule
    \end{tabular}

    \caption{Ablation study for the individual loss term that we presented on~\cref{eq:loss}. The error numbers indicate with full losses, the reconstructed meshes are best in most situation quantitatively.}
    \label{tab:ablation_error}

\end{table}

\newcolumntype{C}[1]{>{\centering\let\newline\\\arraybackslash\hspace{0pt}}m{#1}}
\begin{table}[t]
    \centering
    \scriptsize
    \begin{tabular}{|p{1.2cm}|p{1.4cm}|C{1cm}C{1cm}|C{1cm}C{1cm}|C{1cm}C{1cm}|C{1cm}C{1cm}|}
        \toprule
        \multirow{3}{*}{\textbf{Metric}} & \multirow{3}{*}{\textbf{Dataset}} & \multicolumn{4}{c|}{\textbf{Method} (Sparse)} & \multicolumn{4}{c|}{\textbf{Method} (Dense)} \\
        \cline{3-10}
        & & \multicolumn{2}{c|}{w/o $\mathit{l}_{\mathcal{N}}$} & \multicolumn{2}{c|}{full losses} & \multicolumn{2}{c|}{w/o $\mathit{l}_{\mathcal{N}}$} & \multicolumn{2}{c|}{full losses} \\
        \cline{3-10}
         & & IGR & Ours & IGR & Ours & IGR & Ours & IGR &  Ours\\
        \midrule
        \multirow{4}{*}{\begin{minipage}{1cm}\textbf{CD}\\($\times 10^2$)\end{minipage}} & Bunny & 0.489 & 0.171 & 0.481 & 0.067 & 0.169 & 0.139 & 0.501 & 0.068 \\
         & Armadillo & 0.078 & 0.037 & 0.060 & 0.037 & 0.060 & 0.052 & 0.035 & 0.030 \\
         & Dragon & 0.225 & 0.226 & 0.209 & 0.157 & 0.160 & 0.184 & 0.162  & 0.126\\
         & Buddha & 0.405 & 0.467  & 0.307 & 0.125 & 0.257 & 0.253 & 0.248 & 0.267 \\  
        \midrule
        \multirow{4}{*}{\textbf{HD}} & Bunny & 0.109 & 0.039 & 0.149 & 0.016 & 0.050 & 0.012 & 0.153 & 0.043 \\
        & Armadillo & 0.032 & 0.006 & 0.018 & 0.006 & 0.015 & 0.028 & 0.014 & 0.007 \\
        & Dragon & 0.050 & 0.048 & 0.037 & 0.030 & 0.046 & 0.065 & 0.056 & 0.025 \\
        & Buddha & 0.087 & 0.146 & 0.092 & 0.017 & 0.097 & 0.053 & 0.059 & 0.044 \\
        \bottomrule
    \end{tabular}
    \caption{Quantitative comparison with IGR~\cite{gropp2020implicit} to show the influence of sprase and dense input under no normal information situation.}
    \label{tab:ablation_igr}
\end{table}

\begin{figure}[t!]

    \newcolumntype{Y}{>{\centering\arraybackslash}m{0.12\textwidth}}
    \centering
    \begin{tabular}{YYYYYYYY}
        \multicolumn{2}{c}{Sparse} & \multicolumn{2}{c}{Dense} & \multicolumn{2}{c}{Sparse} & \multicolumn{2}{c}{Dense} \\
\includegraphics[width=\linewidth]{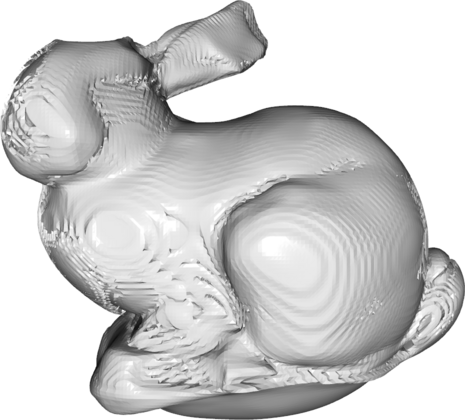} &
\includegraphics[width=\linewidth]{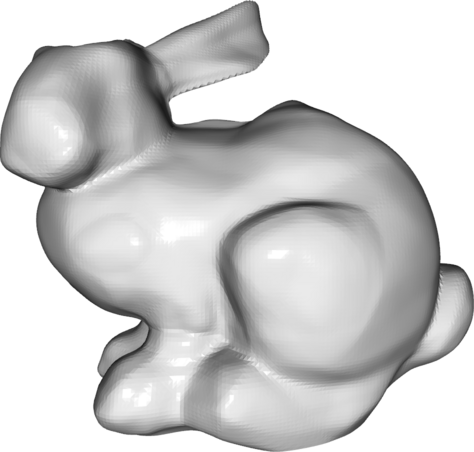} & 
\includegraphics[width=\linewidth]{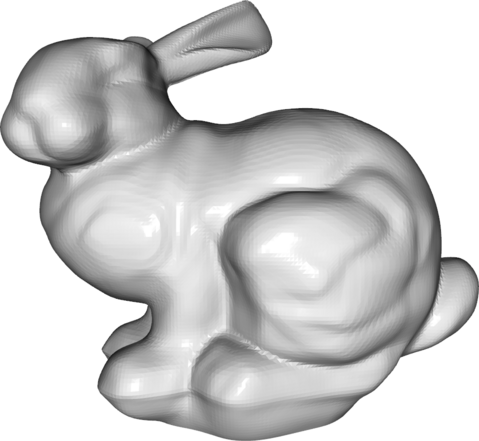} &
\includegraphics[width=\linewidth]{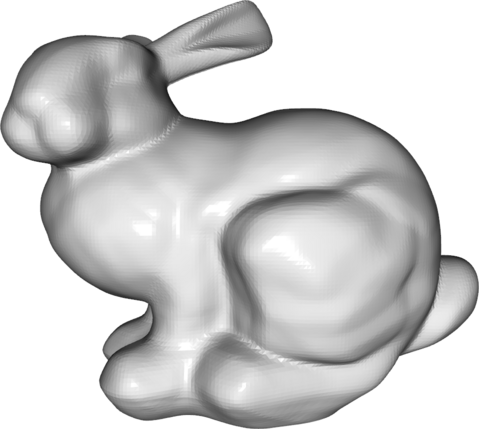} &
\includegraphics[width=\linewidth, clip, viewport=100 0 425 260]{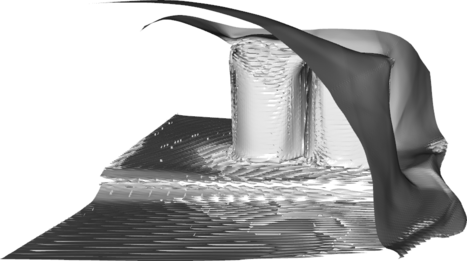} &
\includegraphics[width=\linewidth, clip, viewport=100 0 425 260]{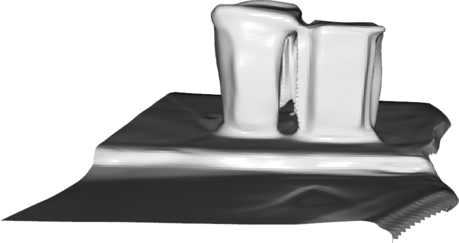} &
\includegraphics[width=\linewidth, clip, viewport=100 0 425 260]{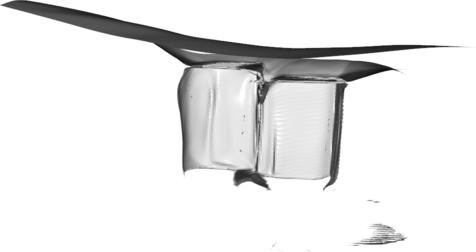} &
\includegraphics[width=\linewidth, clip, viewport=100 0 425 260]{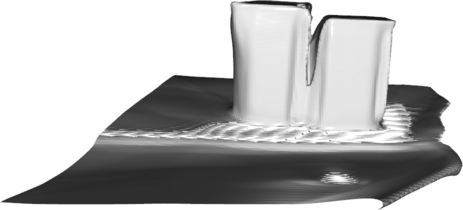}\\
IGR & Ours & IGR & Ours & IGR & Ours & IGR & Ours \\
\includegraphics[width=\linewidth]{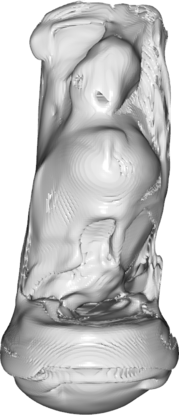} &
\includegraphics[width=\linewidth]{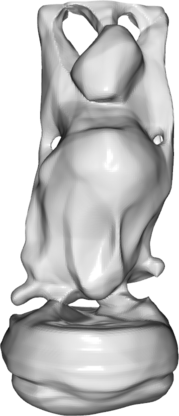} &
\includegraphics[width=\linewidth]{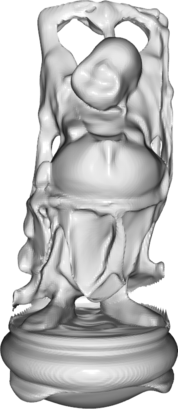} &
\includegraphics[width=\linewidth]{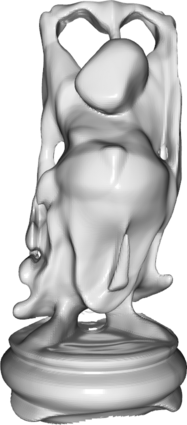} &
\includegraphics[height=\linewidth, origin=c, angle=90, clip, viewport=25 25 400 369]{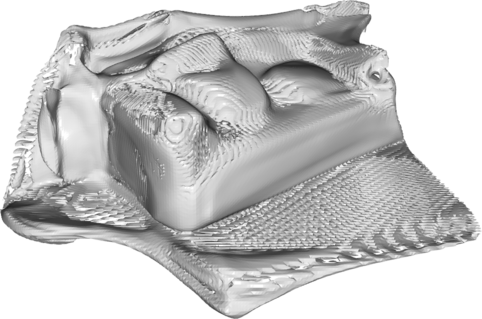}&
\includegraphics[height=\linewidth, origin=c,angle=90, clip, viewport=25 25 400 369]{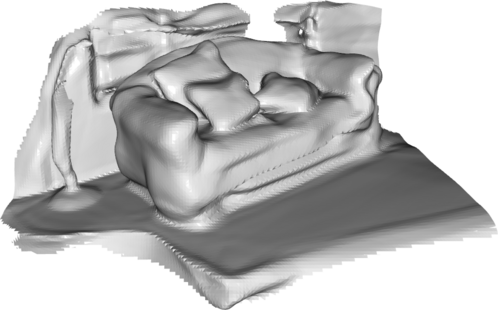} &
\includegraphics[height=\linewidth, origin=c,angle=90, clip, viewport=25 25 400 369]{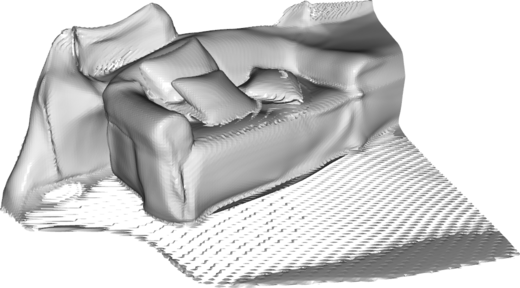} &
\includegraphics[height=\linewidth, origin=c, angle=90, clip, viewport=25 25 400 369]{images/sofa/64/weighted_10000no_color.png}\\
IGR & Ours & IGR & Ours & IGR& Ours & IGR & Ours 
    \end{tabular}
    \caption{No normal information reconstruction results on IGR~\cite{gropp2020implicit} and our method. Our method outperforms IGR on all sparse input situations.}\label{fig:ablation_normal}
    \end{figure}

Even one goal of the proposed method is to solve the normal acquisition problem, especially on real-world datasets. In the ablation study, we still test the influence of off-surface sampling on the reconstruction under the no-normal information case.~\cref{fig:ablation_normal} shows the visualization of two methods, IGR~\cite{gropp2020implicit} and ours, with $\tau_n = 0$ in~\cref{eq:loss}. Our method is robust and stable for sparse input even without normal input. IGR can reconstruct satisfactory meshes for dense situations without normal but it fails when the input is sparse, especially when the surface is planar. Dense input can, therefore, resolve the ambiguity of the surface orientation to some extent. 

\subsection{More Visualization Results}
In this section, we show more visualization results and include the full uncertainty visualization for~\cref{fg:uncertainty} in the main paper. 

\begin{figure}[!ht]
    \centering
        \begin{tabular}{m{0.22\textwidth} m{0.22\textwidth} m{0.22\textwidth} m{0.22\textwidth}}
             \includegraphics*[width=0.2\textwidth]{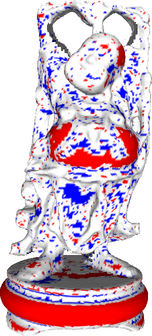} &
            \includegraphics*[width=0.2\textwidth]{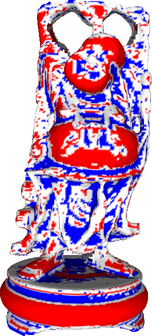}&
             \includegraphics*[height=0.22\textwidth, angle = 90, origin=c]{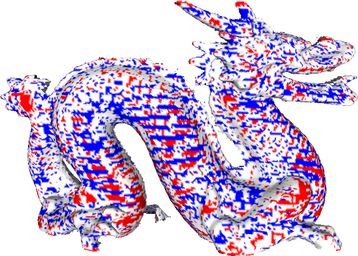} &
             \includegraphics*[height=0.22\textwidth, angle = 90, origin=c]{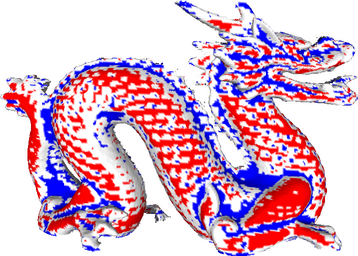} \\
            \centering Gaussian curv & \centering mean curv & \centering Gaussian curv & \centering mean curv
        \end{tabular}
        \caption{The visualization of Gaussian curvatures and mean curvatures of \textit{happy\_buddha} and \textit{dragon} dataset, computed using depth images as described in main paper~\cref{subsec:voxelization}.}\label{fg::curvature2}
        \vspace{-0.5cm}  
\end{figure}

\begin{figure}[t]
    \centering
    \begin{tabular}{m{0.2\linewidth}m{0.2\linewidth}m{0.2\linewidth}m{0.05\linewidth}m{0.2\linewidth}}
        \includegraphics*[width=0.9\linewidth, height=1.6cm]{images/household/64/init_pc_crop.png} &
        \includegraphics*[width=0.9\linewidth, height=1.6cm]{images/household/64/siren_crop.png} &
        \includegraphics*[width=0.9\linewidth, height=1.6cm]{images/household/64/uncertainty_ours2_crop.png} &
        \includegraphics*[height=1.6cm]{images/lr_kt0/64/error_bar_crop.png}&
        \includegraphics*[width=0.9\linewidth, height=1.6cm]{images/household/64/ours4_crop.png}\\
        \includegraphics*[width=0.9\linewidth, height=1.6cm]{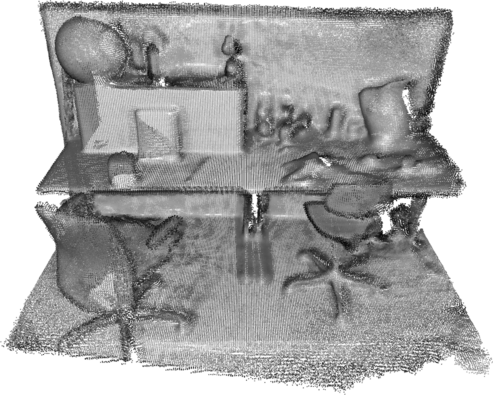} &
        \includegraphics*[width=0.9\linewidth, height=1.6cm]{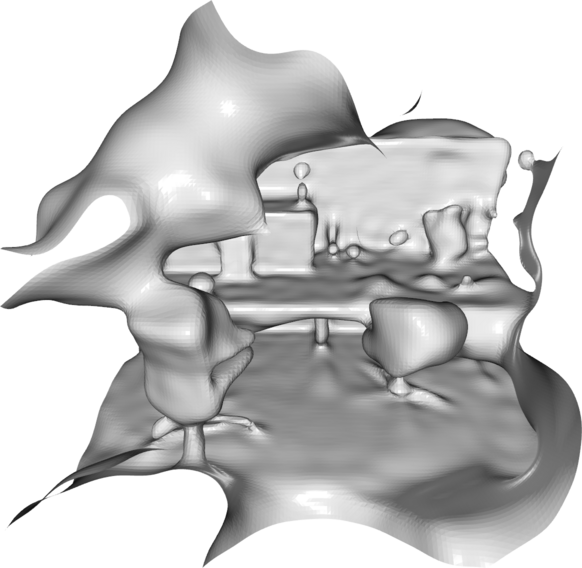} &
        \includegraphics*[width=0.9\linewidth, height=1.6cm]{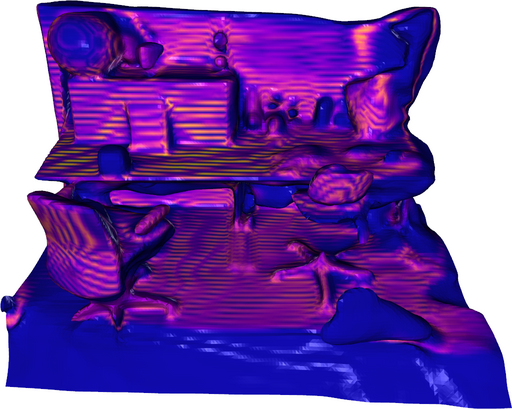} &
        \includegraphics*[height=1.6cm]{images/lr_kt0/64/error_bar_crop.png}&
        \includegraphics*[width=\linewidth]{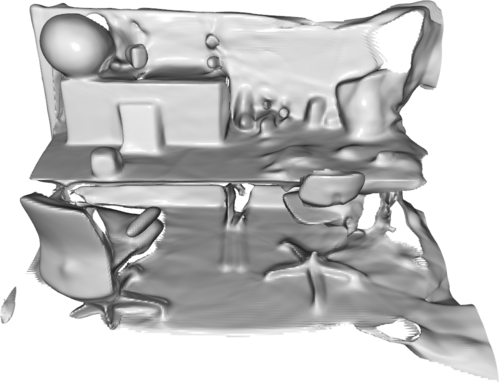}\\
        \centering Input & \centering SIREN & \centering Uncertainty & & {\centering Ours}\\
        \includegraphics*[width=\linewidth, angle=180, origin=c]{images/of_kt0/64/init_pc_crop.png} &
        \includegraphics*[width=\linewidth,angle=180, origin=c]{images/of_kt0/64/igr2_crop.png} &
        \includegraphics*[width=\linewidth,angle=180, origin=c]{images/of_kt0/64/uncertainty_ours_crop.png} &
        \includegraphics*[height=1.5cm]{images/lr_kt0/64/error_bar_crop.png}&
        \includegraphics*[width=\linewidth, angle=180, origin=c]{images/of_kt0/64/ours_crop.png}\\
        \includegraphics*[width=\linewidth, angle=180, origin=c]{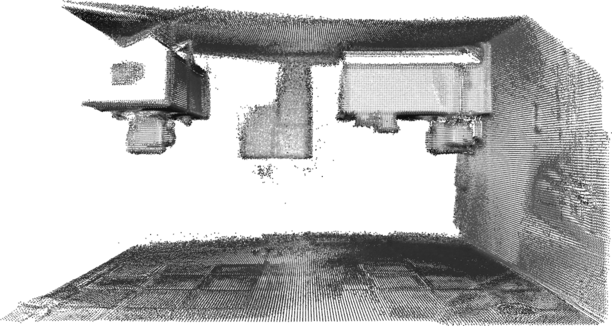} &
        \includegraphics*[width=\linewidth,angle=180, origin=c]{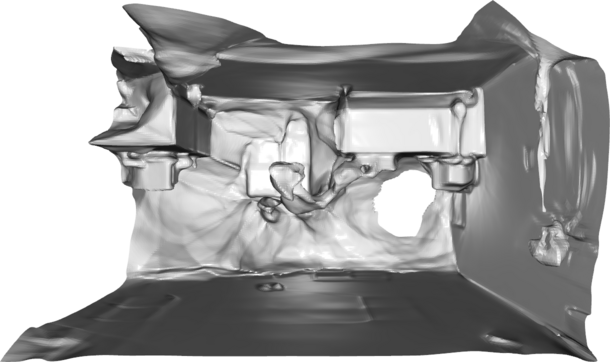} &
        \includegraphics*[width=\linewidth,angle=180, origin=c]{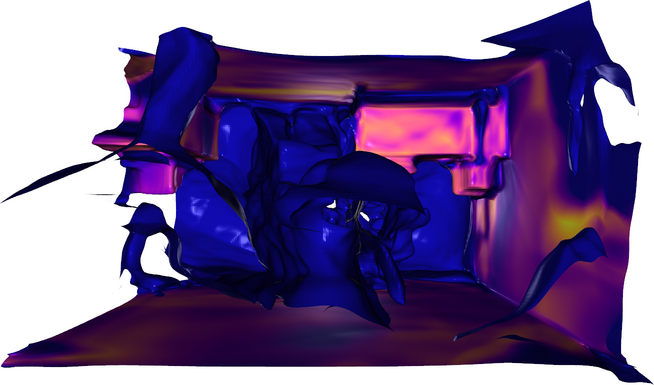} &
        \includegraphics*[height=1.5cm]{images/lr_kt0/64/error_bar_crop.png}&
        \includegraphics*[width=\linewidth, angle=180, origin=c]{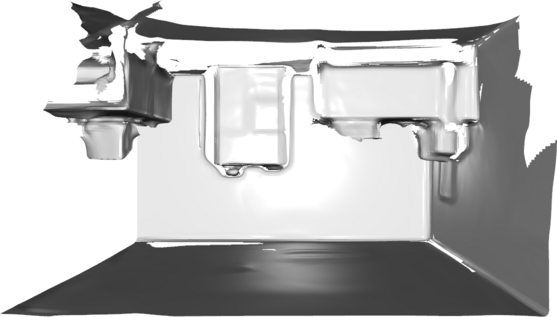}\\
        \centering Input & \centering IGR & \centering Uncertainty & & \centering Ours
    \end{tabular}
    \caption{Scene reconstruction results with sparse input on real-world dataset \emph{TUM\_rgbd} (first rows, sparse points $\sim 14k$, dense points $\sim 330k$) 
    with noisy camera poses. Synthetic dataset \emph{icl\_nium} (last rows, sparse points $\sim 14k$, and dense points $\sim 215k$) 
    with ground truth camera poses.}\label{fg:uncertainty2}
 \vspace*{-0.6cm}
 \end{figure}

\begin{figure}[h!]
    \centering
   \begin{tabular}{m{0.08\linewidth}m{0.08\linewidth}m{0.08\linewidth}m{0.08\linewidth}m{0.08\linewidth}m{0.08\linewidth}m{0.08\linewidth}m{0.08\linewidth}m{0.08\linewidth}}
    \includegraphics*[width=\linewidth]{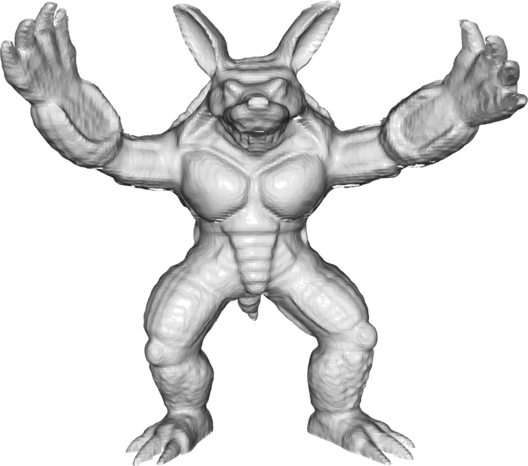} &
    \includegraphics*[width=\linewidth]{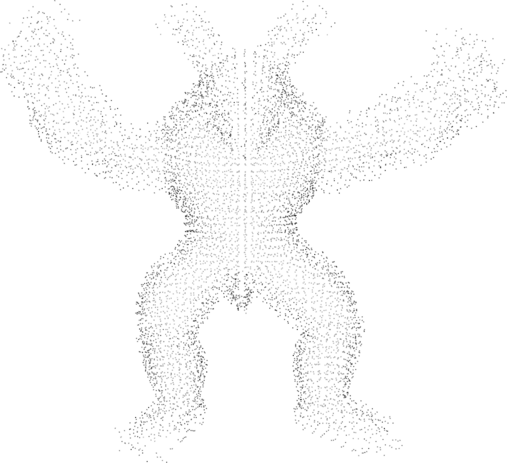} &
    \includegraphics*[width=\linewidth]{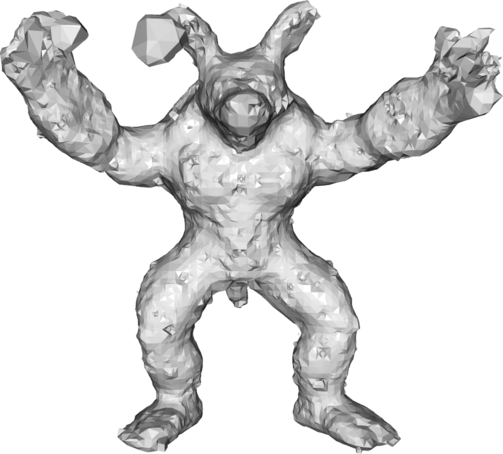} &
    \includegraphics*[width=\linewidth]{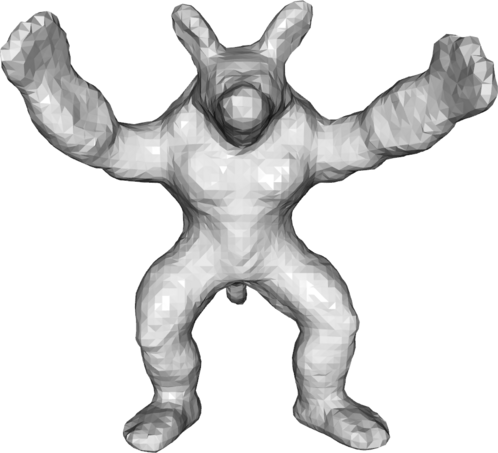}&
    \includegraphics*[width=\linewidth]{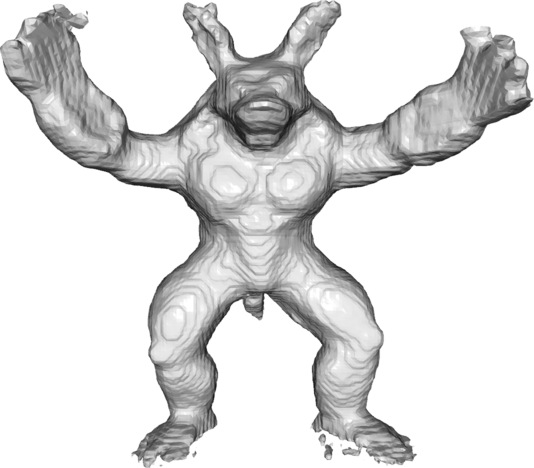} &
    \includegraphics*[width=\linewidth]{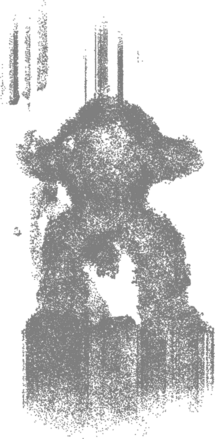} &
    \includegraphics*[width=\linewidth]{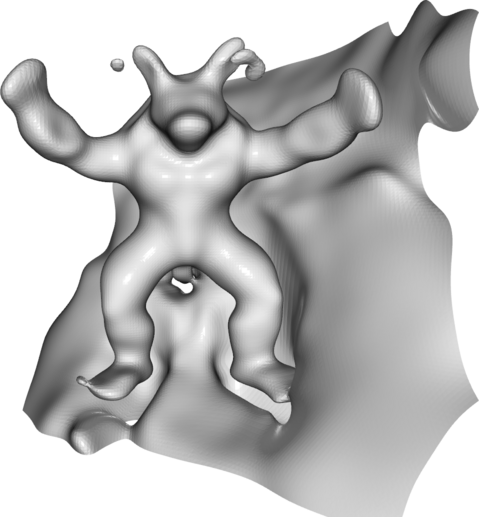} &
    \includegraphics*[width=\linewidth]{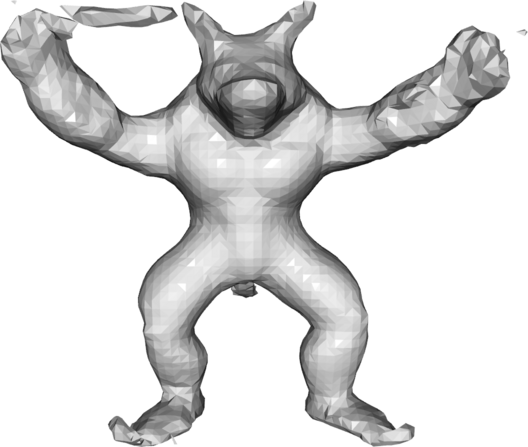} &
    \includegraphics*[width=\linewidth]{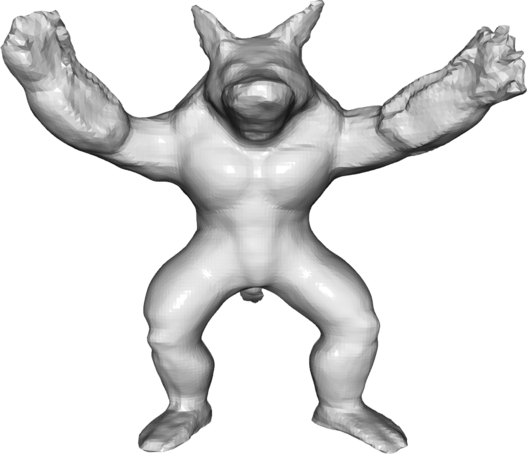} \\
    \includegraphics*[width=\linewidth]{images/armadillo/64/init_mesh_crop.png} &
    \includegraphics*[width=\linewidth]{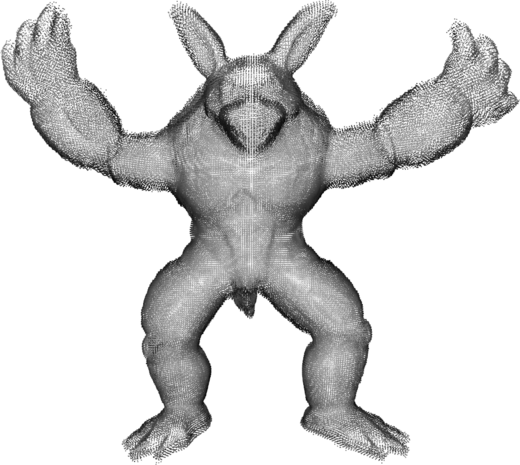} &
    \includegraphics*[width=\linewidth]{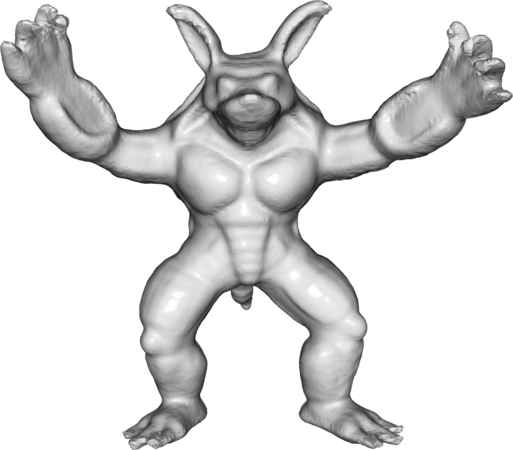} &
    \includegraphics*[width=\linewidth]{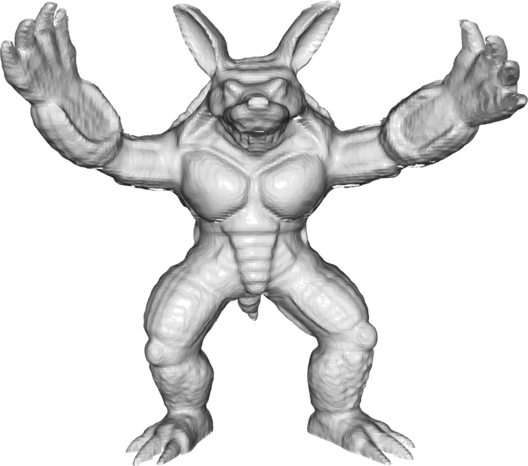}&
    \includegraphics*[width=\linewidth]{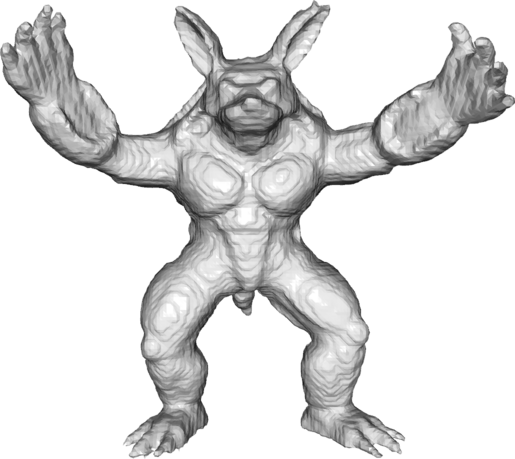} &
    \includegraphics*[width=\linewidth]{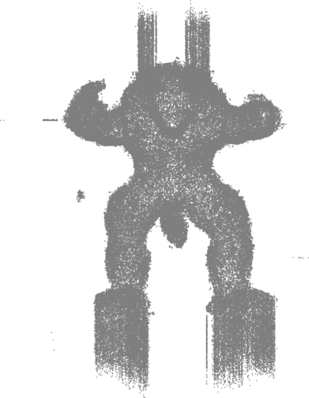}  &
    \includegraphics*[width=\linewidth]{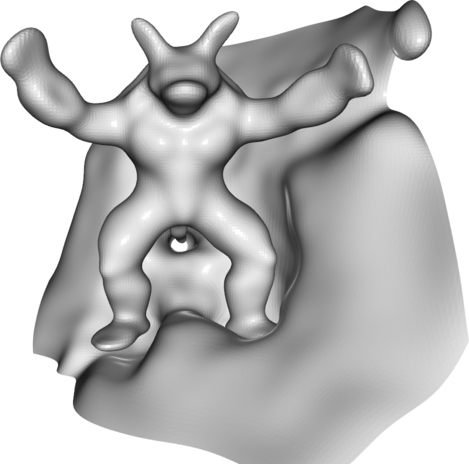} &
    \includegraphics*[width=\linewidth]{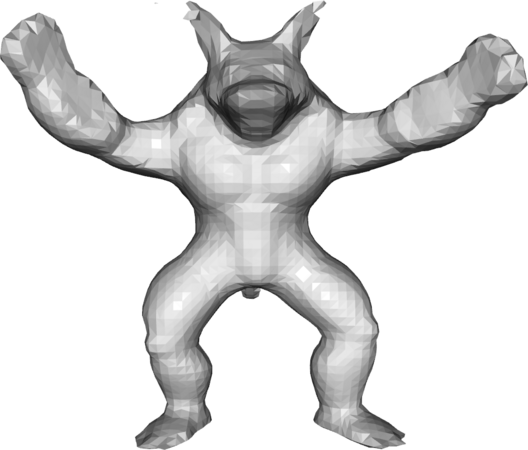} &
    \includegraphics*[width=\linewidth]{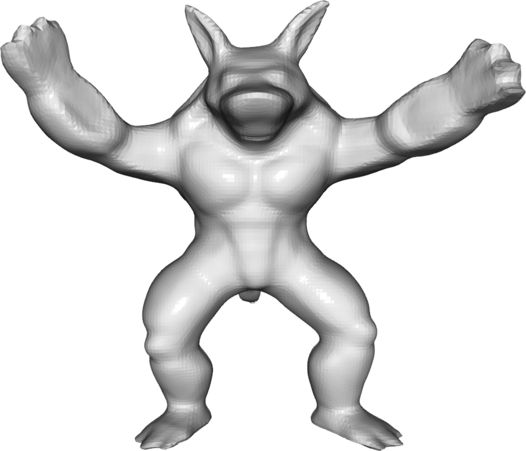} \\
    \includegraphics*[width=\linewidth]{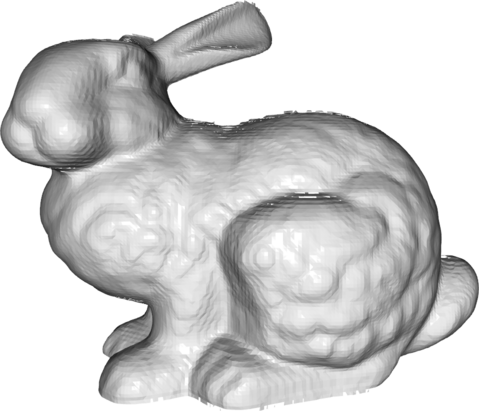} &
    \includegraphics*[width=\linewidth]{images/bunny/64/init_pc_crop.png} &
    \includegraphics*[width=\linewidth]{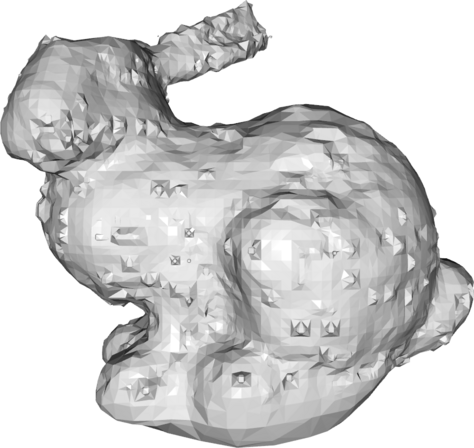} &
    \includegraphics*[width=\linewidth]{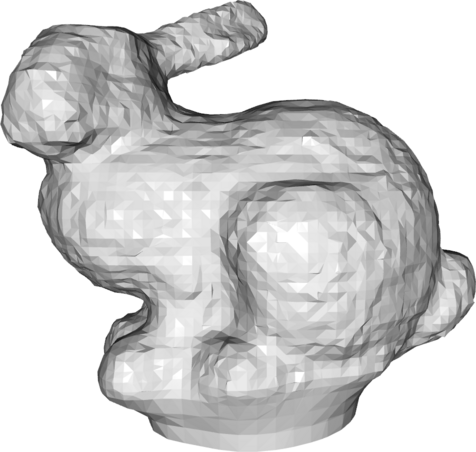}&
    \includegraphics*[width=\linewidth]{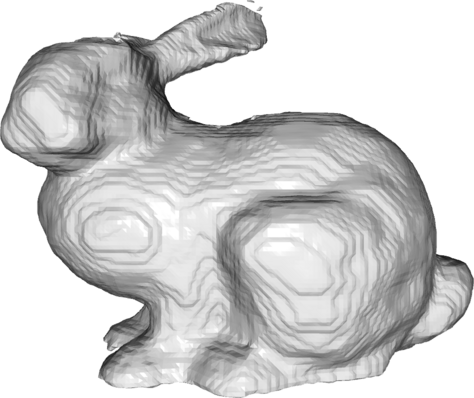} &
    \includegraphics*[width=\linewidth]{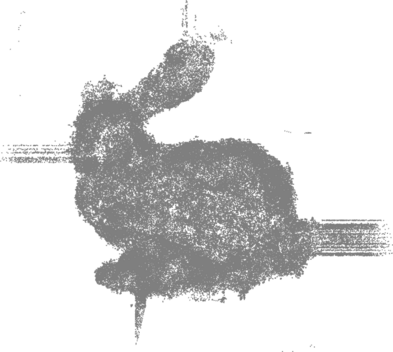}  &
    \includegraphics*[width=\linewidth]{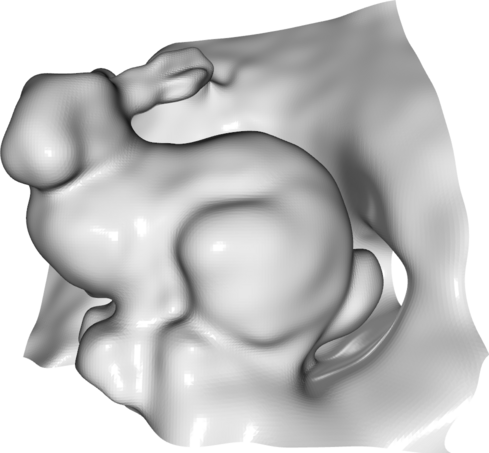} &
    \includegraphics*[width=\linewidth]{images/bunny/64/igr_crop.png} &
    \includegraphics*[width=\linewidth]{images/bunny/64/ours_igr_crop.png} \\
    \includegraphics*[width=\linewidth]{images/bunny/256/init_mesh_crop.png} &
    \includegraphics*[width=\linewidth]{images/bunny/256/init_pc_crop.png} &
    \includegraphics*[width=\linewidth]{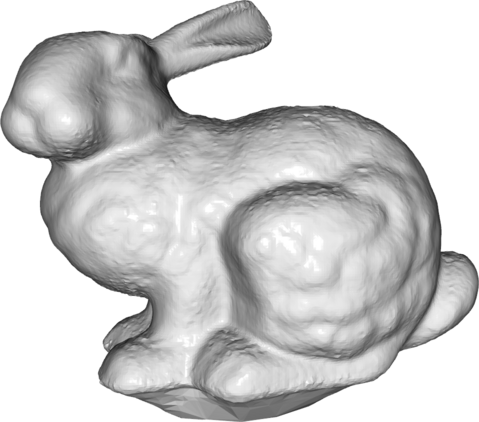} &
    \includegraphics*[width=\linewidth]{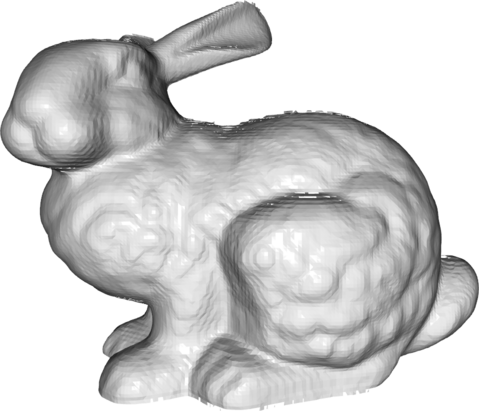}&
    \includegraphics*[width=\linewidth]{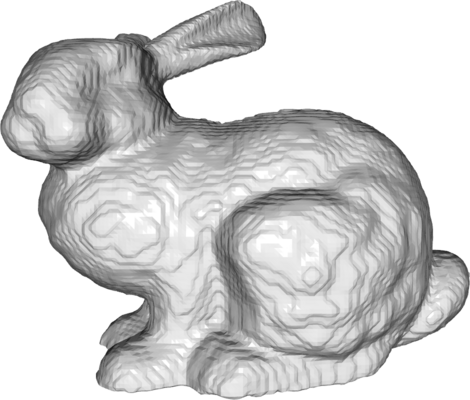} &
    \includegraphics*[width=\linewidth]{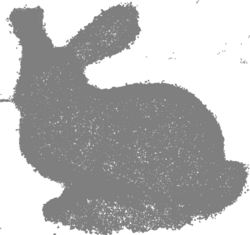}  &
    \includegraphics*[width=\linewidth]{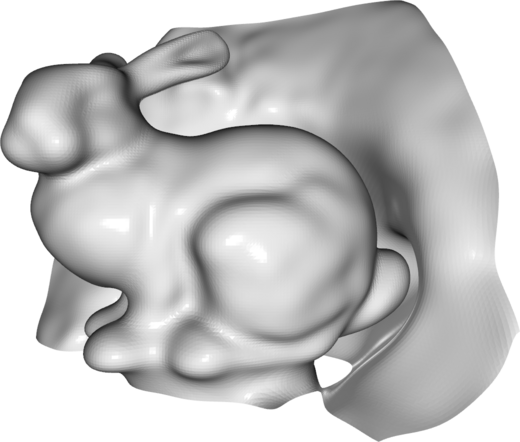} &
    \includegraphics*[width=\linewidth]{images/bunny/256/igr_crop.png} &
    \includegraphics*[width=\linewidth]{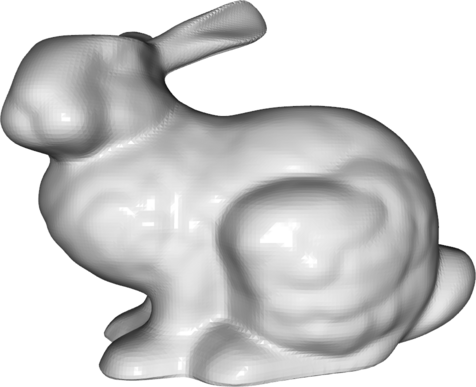} \\
    
   \centering GT mesh & \centering Input & \centering SSD & \centering Poisson & \centering IF-NET & \centering NDF & \centering SIREN & \centering IGR & {\centering Ours}
   \end{tabular}
   \caption{Comparison results with SSD~\cite{Calakli2011}, Poisson surface recontruction~\cite{Kazhdan2006poisson}, NDF~\cite{chibane2020ndf}, IGR~\cite{gropp2020implicit}, SIREN~\cite{sitzmann2019siren} and IF-NET~\cite{chibane2020implicit} with two different density input on synthetic datasets~\cite{bunny}. \emph{Armadillo} has $\sim 8k$ in sparse input and $\sim 146k$ in dense input. \emph{Bunny} has $\sim 5k$ sparse points and dense one with $\sim 100k$ points}\label{fg:appendix_visual}
\end{figure}

\begin{figure}[h!]
    \centering
   \begin{tabular}{m{0.13\linewidth}m{0.13\linewidth}m{0.13\linewidth}m{0.13\linewidth}m{0.13\linewidth}m{0.02\linewidth}m{0.13\linewidth}}
    \includegraphics*[width=\linewidth]{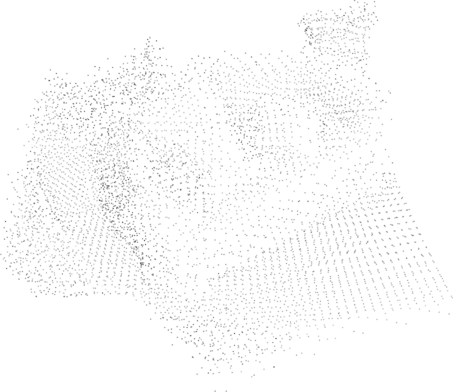}&
    \includegraphics*[width=\linewidth]{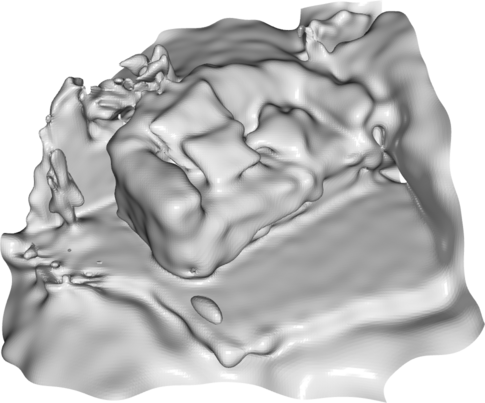} &
    \includegraphics*[width=\linewidth]{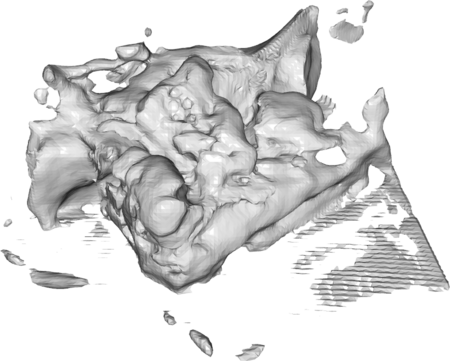}&
    \includegraphics*[width=\linewidth]{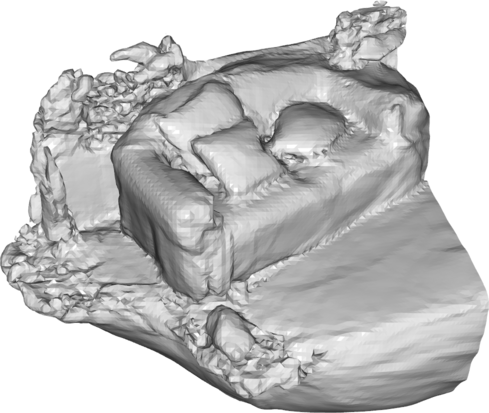}&
    \includegraphics*[width=\linewidth]{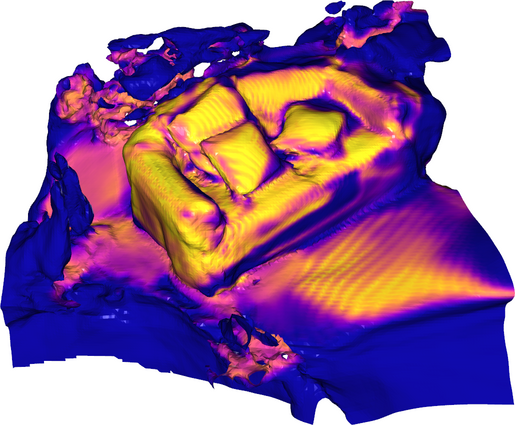}&
    \includegraphics*[height=1.7cm]{images/lr_kt0/64/error_bar_crop.png}&
    \includegraphics*[width=\linewidth]{images/sofa/64/ours2_crop.png} \\
    \includegraphics*[width=\linewidth]{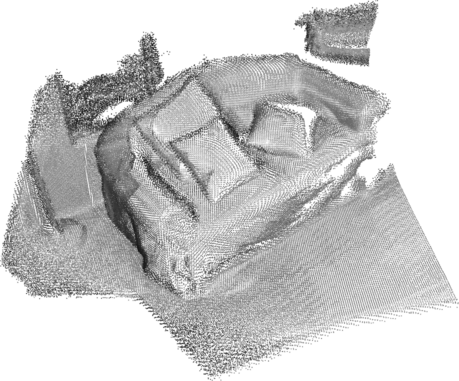}&
    \includegraphics*[width=\linewidth]{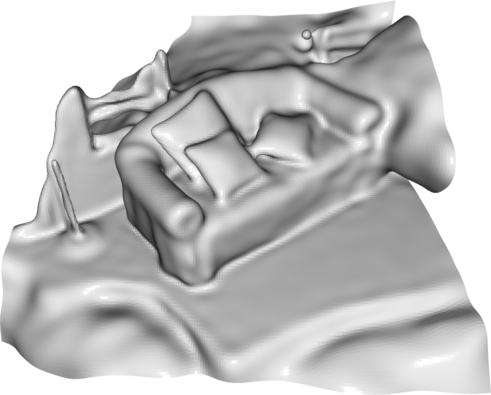} &
    \includegraphics*[width=\linewidth]{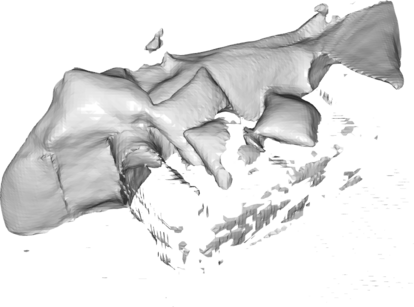}&
    \includegraphics*[width=\linewidth]{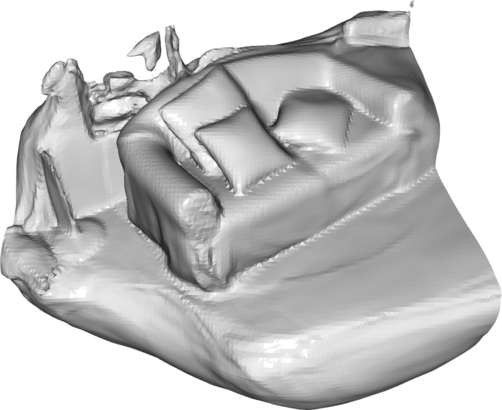}&
    \includegraphics*[width=\linewidth]{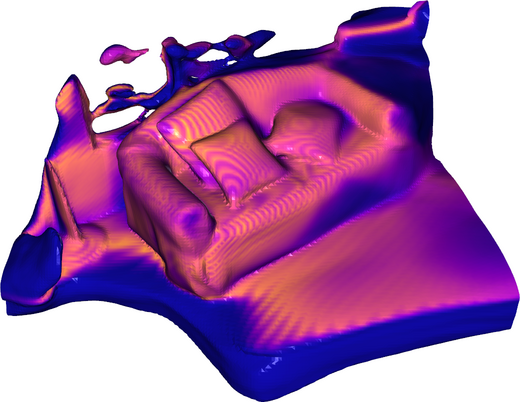}&
    \includegraphics*[height=1.7cm]{images/lr_kt0/64/error_bar_crop.png}&
    \includegraphics*[width=\linewidth]{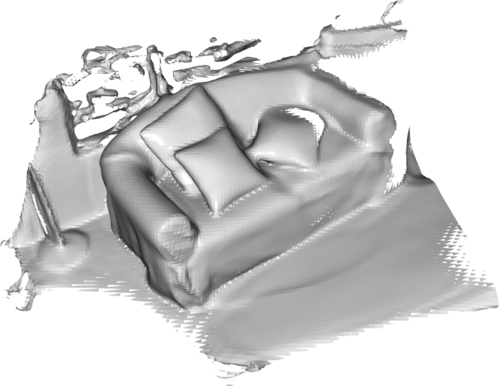}\\
    \includegraphics*[width=\linewidth]{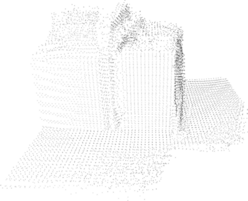}&
    \includegraphics*[width=\linewidth]{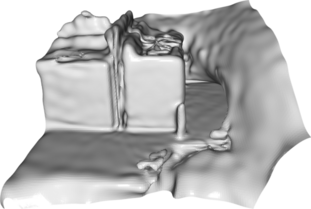} &
    \includegraphics*[width=\linewidth]{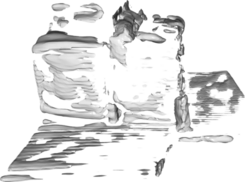}&
    \includegraphics*[width=\linewidth]{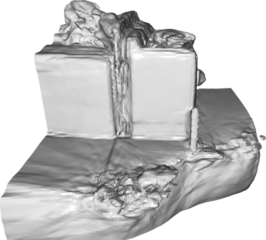}&
    \includegraphics*[width=\linewidth]{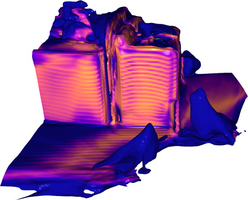}&
    \includegraphics*[height=1.7cm]{images/lr_kt0/64/error_bar_crop.png}&
    \includegraphics*[width=\linewidth]{images/washmachine/64/ours_crop.png} \\
    \includegraphics*[width=\linewidth]{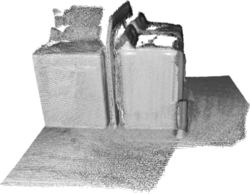}&
    \includegraphics*[width=\linewidth]{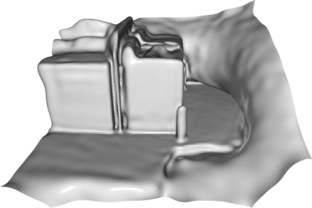} &
    \includegraphics*[width=\linewidth]{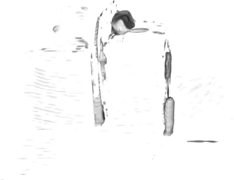}&
    \includegraphics*[width=\linewidth]{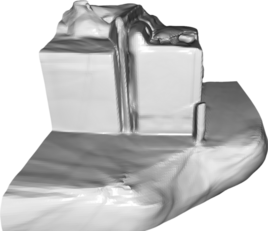}&
    \includegraphics*[width=\linewidth]{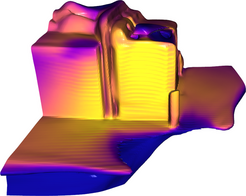}&
    \includegraphics*[height=1.7cm]{images/lr_kt0/64/error_bar_crop.png}&
    \includegraphics*[width=\linewidth]{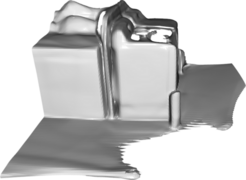}\\
    \centering Input & \centering SIREN & NeuralPull & \centering IGR & \centering Uncertainty& & \centering Ours
   \end{tabular}
   \caption{Two \textbf{Real world} Scene datasets~\cite{redwood} results comparison with SIREN~\cite{sitzmann2019siren}, NeuralPull~\cite{ma2020neuralpull} and IGR~\cite{gropp2020implicit}. NeuralPull fails in most of the cases, while SIREN creates a redundant area. \emph{Sofa} has $\sim 6k$ points and $\sim 111k$ points in sparse and dense situation, respectively. \emph{Washmachine} with $\sim 8k$ sparse points and $\sim 180k$ dense points.}
   \end{figure}

   \section{Failing Cases Analysis}
   
   \begin{figure}[h!]
    \centering
   \begin{tabular}{m{0.18\linewidth}m{0.18\linewidth}m{0.18\linewidth}m{0.18\linewidth}m{0.18\linewidth}}
    \includegraphics*[width=\linewidth, angle=180, origin=c]{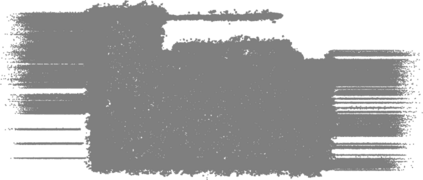} &
    \includegraphics*[width=\linewidth, angle=180, origin=c]{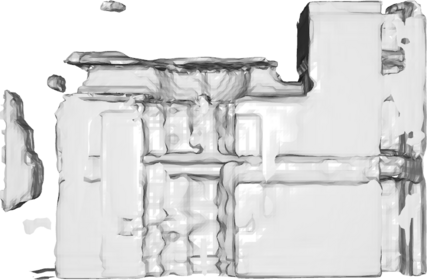}&
    \includegraphics*[width=\linewidth, angle=180, origin=c]{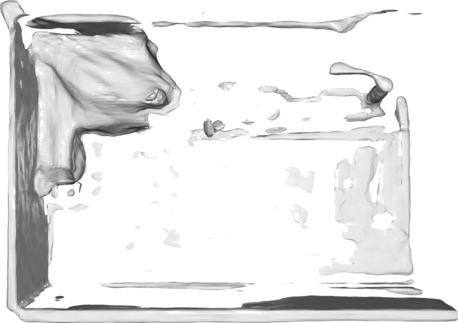}&
    \includegraphics*[width=\linewidth, angle=180, origin=c]{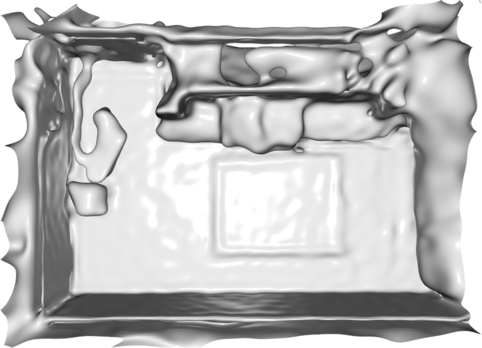}&
    \includegraphics*[width=\linewidth, angle=180, origin=c]{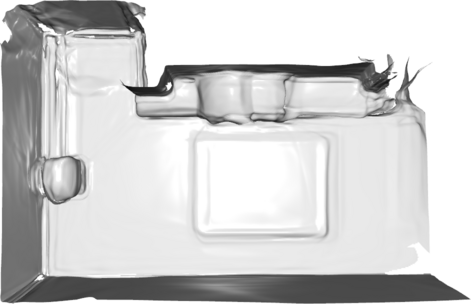} \\
    \centering NDF & \centering IF-NET & \centering NeuralPull & \centering SIREN & {\centering Ours} \\
    \includegraphics*[width=\linewidth]{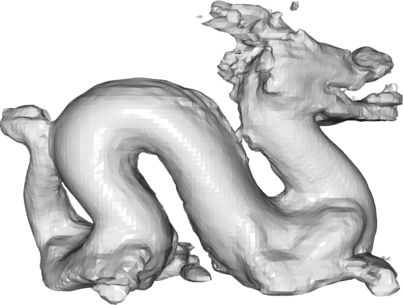} &
    \includegraphics*[width=\linewidth]{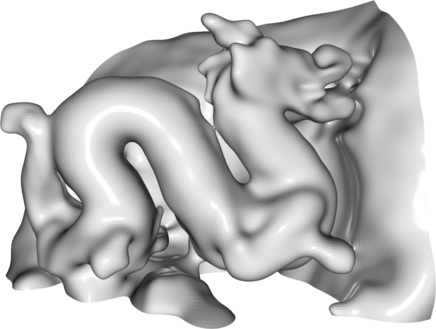} &
    \includegraphics*[width=\linewidth]{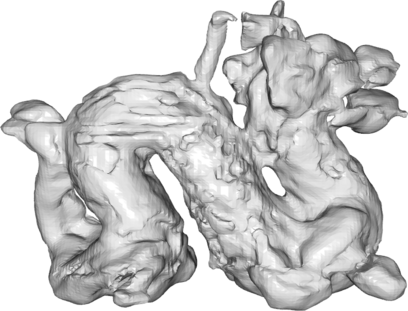} &
    \includegraphics*[width=\linewidth]{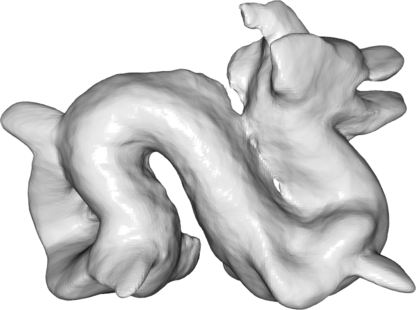} &
      \includegraphics*[width=\linewidth]{images/dragon/64/ours_igr_crop.png} \\
      \centering IGR & \centering SIREN & \centering NeuralPull & \centering Modified NeuralPull & \centering Ours
   \end{tabular}
   \caption{Failing cases in different methods.}\label{fg:failing_case}
   \end{figure}

   Chibane \etal show in~\cite{chibane2020ndf} successful results in ShapeNet~\cite{chang2015shapenet}. However, we did not get satisfactory results on both object and scene datasets. We suspect the method needs a lot of training data for one single shape, \eg, ShapeNet has multiple point clouds for a single shape. We only have one (sparse) point cloud for a shape. We also see a similar issue reported in the authors' Github issues. \cref{fg:failing_case} shows different failing cases. The first line is the failing case on open surface reconstruction for sparse input ($\sim 6k$) \emph{lr\_kt0} datasets. IF-NET~\cite{chibane2020implicit} can not handle open surfaces. NeuralPull~\cite{ma2020neuralpull} easily fails when there is a flat plane. One reason is NeuralPull learns to pull the sampled point to the closest surface point. If the sampled point is already on the surface, it learns to pull back and forth on the same plane. SIREN~\cite{sitzmann2019siren} tends to create artifices in non-surface areas when dealing with open surfaces as it use periodic activation functions after each layer, it creates random area when initialize the neural network. The second row is failing cases for complicated shape \emph{Dragon} (sparse input, $\sim 6k$ points). We also fail to recover satisfactory results using our method (last column), and the modified NeuralPull method with our sampling strategy also fails to recover the correct shape of the Dragon (last second column). Our analysis is that large details are gathered around the head part of the Dragon, and the interpolation fails to overcome too sparse input.

\section{Code, Datasets, and Baseline Methods}

 The following table contains detailed information (link, license) of used code and datasets in the main paper.

\begin{longtable}{lp{1.5cm}ll p{5cm} p{2.5cm}}
    \toprule
         & name & type & year & link & license \\
    \midrule
        \cite{redwood} & Redwood & dataset & 2016 & {\small\url{http://www.redwood-data.org/3dscan/}} & Public Domain \\
        \cite{bunny} & The Stanford 3D & dataset & 1994 & {\small\url{http://graphics.stanford.edu/data/3Dscanrep/}} & Public Domain \\
        \cite{zollhofer2015shading} & multi-view dataset & dataset & 2015 & {\small\url{http://graphics.stanford.edu/projects/vsfs/}} & CC BY-NC-SA 4.0 \\
       \cite{iclnium} & ICL-NUIM & dataset & 2014 & {\small\url{https://www.doc.ic.ac.uk/~ahanda/VaFRIC/iclnuim.html}} & CC BY 3.0 \\
        \cite{sturm12iros} & TUM-rgbd & dataset & 2012 & {\small\url{https://cvg.cit.tum.de/data/datasets/rgbd-dataset}} & CC BY 4.0 \\
        \cite{sommer2022} & gradient-SDF & code & 2022 & {\small\url{https://github.com/c-sommer/gradient-sdf}} & BSD-3 \\
        \cite{gropp2020implicit} & IGR & code & 2020 & {\small\url{https://github.com/amosgropp/IGR}} & - \\
        \cite{weder2020routfusion} & Routed-Fusion & code & 2020 & {\small\url{https://github.com/weders/RoutedFusion}} & - \\
        \cite{sitzmann2019siren} & SIREN & code & 2019 & {\small\url{https://github.com/vsitzmann/siren}} & MIT license \\
        \cite{chibane2020implicit} &IF-NET & code & 2020 & {\small\url{https://virtualhumans.mpi-inf.mpg.de/ifnets/}} & - \\
        \cite{ma2020neuralpull} & NeuralPull & code & 2021 & {\small\url{https://github.com/bearprin/neuralpull-pytorch}} & - \\
        \cite{chibane2020ndf} & NDF & code & 2020 & {\small\url{https://virtualhumans.mpi-inf.mpg.de/ndf/}} & -\\
        \cite{Kazhdan2006poisson} & Poisson & code & 2006 & {\small\url{http://www.open3d.org/}} & - \\
        \cite{Calakli2011} & SSD & code & 2011 & {\small\url{http://mesh.brown.edu/ssd/software.html}} & - \\
    \bottomrule \\
  \caption{Used datasets and code in our submission, with reference, link, and license. We did our real-world experiments on two datasets, multi-view dataset~\cite{zollhofer2015shading} (for which ground truth poses exist), and Redwood~\cite{redwood} (without ground truth poses). There are two synthetic datasets: the Stanford 3D~\cite{bunny}, an object dataset, and the ICL-NUIM dataset~\cite{iclnium}, a scene dataset. For the comparison methods, we use the code listed in the table.}    \label{tab:code_data}

\end{longtable}
\end{document}